\documentclass [runningheads] {llncs}

\usepackage [T1]               {fontenc}
\usepackage [utf8]             {inputenc}
\usepackage [hidelinks]        {hyperref}
\usepackage [scale = .8]       {noto-mono}
\usepackage [colorinlistoftodos,
             prependcaption,
             textsize = tiny]  {todonotes}
\usepackage [noend]            {algpseudocode}

\usepackage{
  algorithm,
  amsfonts,
  amsmath,
  amssymb,
  array,
  booktabs,
  bm,
  cleveref,
  comment,
  cite,
  forest,
  graphicx,
  listings,
  mathtools,
  microtype,
  multirow,
  nicefrac,
  orcidlink,
  pgfplots,
  soul,
  subfigure,
  tabularx,
  tikz,
  url,
  wrapfig,
  xcolor,
  xspace,
}

\pgfplotsset {compat = 1.18}

\usetikzlibrary{spy}

\definecolor {codegreen}  {rgb} {0,0.6,0}
\definecolor {codegray}   {rgb} {0.5,0.5,0.5}
\definecolor {codepurple} {rgb} {0.58,0,0.82}
\definecolor {backcolour} {rgb} {0.95,0.95,0.92}
\definecolor {pastelblue} {rgb} {0.68, 0.78, 0.81}

\crefname {algorithm}   {Alg.}        {algorithms}
\Crefname {algorithm}   {Algorithm}   {Algortihm}
\crefname {appendix}    {Appendix}    {appendices}
\crefname {equation}    {Eq.}         {equations}
\Crefname {equation}    {Equation}    {Equations}
\crefname {proposition} {Prop.}       {propositions}
\Crefname {proposition} {Proposition} {Propositions}
\crefname {lemma}       {Lemma}       {lemmata}
\Crefname {lemma}       {Lemma}       {Lemmata}
\crefname {listing}     {Listing}     {listings}
\Crefname {listing}     {Listing}     {Listings}
\crefname {definition}  {Def.}        {definitions}
\Crefname {definition}  {Definition}  {Definitions}
\crefname {theorem}     {Thm.}        {theorems}
\Crefname {theorem}     {Theorem}     {Theorems}
\crefname {figure}      {Fig.}        {figures}
\Crefname {figure}      {Figure}      {Figures}
\crefname {page}        {p.}          {pages}
\Crefname {page}        {Page}        {Pages}
\crefname {section}     {Sect.}       {sections}
\Crefname {section}     {Section}     {Sections}
\crefname {example*}    {Example}     {examples}
\Crefname {example*}    {Example}     {Examples}
\crefname {table}       {Tbl.}        {tables}
\Crefname {table}       {Table}       {Tables}

\lstdefinestyle {mystyle} {
  language = Python,
  basicstyle = \ttfamily\footnotesize,
  keywordstyle = \color{blue},
  commentstyle = \color{green},
  numbers = left,
  numberstyle = \sffamily\tiny\color{gray},
  breaklines = true,
  breakatwhitespace = false,
  showtabs = false,
  tabsize = 2,
  keepspaces = true,
  captionpos = b,
  numbersep = 5pt,
  literate = {\ \ }{{\ }}1,
  columns = fullflexible,
}

\lstdefinestyle {myJavastyle} {
    language = Java,
    basicstyle = \ttfamily\footnotesize,
    keywordstyle = \color{orange},
    commentstyle = \color{blue},
    numbers = left,
    numberstyle = \tiny\color{gray},
    breaklines = true,
    breakatwhitespace = false,
    showtabs = false,
    tabsize = 2,
    keepspaces = true,
    captionpos = b,
    numbersep = 5pt,
    literate = {\ \ }{{\ }}1,
    columns = fullflexible,
}

\lstdefinelanguage{XML}
{
  morestring=[b]",
  morecomment=[s]{<!--}{-->},
  moredelim=[s][\color{blue}]{<}{>},
  moredelim=[s][\color{red}]{</}{>},
  moredelim=[l][\color{red}]{/>},
  moredelim=[l][\color{red}]{>},
  stringstyle=\color{black},
  identifierstyle=\color{black},
  keywordstyle=\color{blue},
  morekeywords={xmlns,version,type}
}

\lstnewenvironment{xml}[1][]
{
  \lstset{
    language=XML,
    basicstyle=\ttfamily,
    columns=fullflexible,
    keywordstyle=\color{blue}\bfseries,
    stringstyle=\color{red},
    commentstyle=\color{gray}\textit,
    breaklines=true,
    showstringspaces=false,
    captionpos=b,
    #1
  }
}
{}

\makeatletter
\newcommand \optSub [1]%
  {\ifx #1 \empty \else _{#1} \fi}
\makeatother

\newcommand \generalspace
  {-2mm}

\newcommand \RND
  {\ensuremath {\textrm {rnd}}\xspace}
\newcommand \True
  {\ensuremath {\textrm {True}}\xspace}
\newcommand \False
  {\ensuremath {\textrm {False}}\xspace}
\newcommand \Ops
  {\ensuremath {\textrm {Ops}}\xspace}

\newcommand \StateSet
  {\ensuremath {\mathcal S}\xspace}

\newcommand \EnvState [1] []
  {\ensuremath {\overline { \mathcal S}\optSub{#1}}}

\newcommand \IniState
  {\ensuremath {\EnvState[0]}\xspace}

\newcommand \Action
  {\ensuremath {\mathcal A}\xspace}

\newcommand \observe
  {\ensuremath {\mathcal O}\xspace}

\newcommand \step
  {\ensuremath {\mathcal T}\xspace}

\newcommand \reward
  {\ensuremath {\mathcal R}\xspace}

\newcommand \isFinal
  {\ensuremath {\mathcal F}\xspace}

\newcommand \Pcal
  {\ensuremath {\mathcal P}\xspace}

\newcommand \EnvS [1] []
  {\ensuremath {\overline s}\xspace}

\newcommand \Env
  {\ensuremath {\mathit {Env}}}

\newcommand \PC
  {\ensuremath {\mathit {PC}}\xspace}

\newcommand \RR     {\ensuremath {\mathbb R}}

\newcommand \PDF [1]
  {\textnormal {pdf } #1\xspace}

\renewcommand{\epsilon}{\varepsilon}

\newtheorem {example*} {Example}

\newcolumntype{R}[2]{%
    >{\adjustbox{angle=#1,lap=\width-(#2)}\bgroup}%
    l%
    <{\egroup}%
}

\newcommand \partitioning
{partitioning}

\newcommand \Partitioning
{Partitioning}

\newcommand \partition
{partition}

\newcommand \partitions
{partitions}

\newcommand \Partition
{Partition}

\newcommand \Partitions
{Partitions}

\newcommand \parti
{part}

\newcommand \parts
{parts}

\begin {document}

\title {Symbolic State Partitioning\\ for Reinforcement Learning}


 \author{
 Mohsen Ghaffari\inst{1}\,\orcidlink{0000-0002-1939-9053}\and
 Mahsa Varshosaz\inst{1}\,\orcidlink{0000-0002-4776-883X}\and\\
 Einar Broch Johnsen\inst{2}\,\orcidlink{0000-0001-5382-3949}\and
 Andrzej Wąsowski\inst{1}\,\orcidlink{0000-0003-0532-2685}}
 \institute{
 ITU, Copenhagen, Denmark\\
 \email{\{mohg, mahv, wasowski\}@itu.dk} \and
 University of Oslo, Oslo, Norway\\
 \email{einarj@ifi.uio.no}}

 \authorrunning{M.~Ghaffari et al.}

\maketitle

\begin {abstract}
Tabular reinforcement learning methods cannot operate directly on continuous state spaces. One solution to this problem is to \partition\ the state space. A good \partitioning\ enables generalization during learning and more efficient exploitation of prior experiences. Consequently, the learning process becomes faster and  produces more reliable policies. However, \partitioning\ introduces approximation, which is particularly harmful in the presence of nonlinear relations between state components. An ideal \partition\ should be as coarse as possible, while capturing the key structure of the state space for the given problem. This work extracts \partitions\ from the environment dynamics by symbolic execution. We show that symbolic \partitioning\ improves state space coverage with respect to environmental behavior and allows reinforcement learning to perform better for sparse rewards. We evaluate symbolic state space \partitioning\  with respect to precision, scalability, learning agent performance and state space coverage for the learned policies.\looseness -1

\keywords {Reinforcement Learning  \and Symbolic Execution \and State Space Partitioning}
\end{abstract}

\section {Introduction}%
\label {sec:introduction}

\emph {Reinforcement learning} is a form of active learning, where an agent learns to make decisions to maximize a reward signal.  The agent interacts with an environment and takes actions based on its current state.  The environment rewards the agent, which uses the reward value to update its decision-making policy (see \cref{fig:rl}).  Reinforcement learning has applications in many domains: robotics\,\cite {kober2013reinforcement}, gaming\,\cite {szita2012reinforcement}, electronics\,\cite {ghaffari2021learning}, and healthcare\,\cite {yu2021reinforcement}.  This method can automatically synthesize controllers for many challenging control problems\,\cite {sutton.barto:2018}; however, dedicated approximation techniques, hereunder deep learning, are needed for continuous state spaces.  Unfortunately, despite many successes with continuous problems, Deep Reinforcement Learning suffers from low explainability and lack of convergence guarantees.   At the same time, discrete (tabular) learning methods have been shown to be more explainable\,\cite {Vyetrenko2019,Madumal2020,puiutta2020explainable,zelvelder2021assessing} and to yield policies for which it is easier to assure safety\,\cite {fulton2018safe,VERDIER2019230,jansson2023discretization}, for instance using formal verification\,\cite {jin2022cegar,tran2019safety,adelt2022towards}. Thus, finding a good state space representation for discrete learning remains an active research area\,\cite {peter2019partitioning,lee2004adaptive,wei2018q,akrour2018regularizing,mavridis2021vector,nicol2012states,pmlr-v216-dadvar23a}.
\looseness -1

To adapt a continuous state space for discrete learning, one exploits partial observability to merge regions of the state space into discrete \partitions. Each \parti\ in a partition represents a subset of the states of the agent. Ideally, all states in a \parti\ capture meaningful aspects of the environment---best if they share the same optimal action in the optimal policy.  Consequently, a good \partitioning\ is highly problem specific. For instance, in safety critical environments, it is essential to identify small ``singularities''---regions that require special handling---even if they are very small.  Otherwise, if such regions are included in a larger \parti, the control policy will not be able to distinguish them from the surrounding \parts, leading to high variance in operation time and slow convergence of learning.
\looseness -1

The trade-off between the size of the \partitions\ and the optimality and convergence of reinforcement learning remains a challenge\,\cite {lee2004adaptive,wei2018q,akrour2018regularizing,mavridis2021vector,nicol2012states,pmlr-v216-dadvar23a}. Policies obtained for coarse \partitions\ are unreliable. Large fine \partitions\ make reinforcement learning slow. The dominant methods are \emph{tiling} and \emph {vector quantization}\,\cite {lee2004adaptive,wei2018q,mavridis2021vector,nicol2012states}; neither is adaptive to the structure of the state space. They ignore nonlinear dependencies between state components even though quadratic behaviors are common in control systems.  So far, the shape of the state space \partitions\ has hardly been studied in the literature.

In this work, we investigate the use of \emph{symbolic execution} to extract approximate adaptive \partitions\ that reflect the problem dynamics. \emph{Symbolic execution}\,\cite{king1976symbolic,clarke1976program} is a classic foundational technique for dynamic program analysis, originating in software engineering and deductive verification research and commonly used for test input generation\,\cite{visser.ea:2006:test} and in interactive theorem provers (e.g.\,\cite{ahrendt.ea:2016:deductive}). A symbolic executor generates a set of \emph{path conditions} (\PC), constraints that must hold for each execution path that the program can take.  These conditions \partition\ the state space of the executed program into groups that share the same execution path. Our hypothesis is that \emph {the path conditions obtained by symbolic execution of an environment model (the step and reward functions) provide a useful state space \partition\ for reinforcement learning}.  The branches in the environment program likely reflect important aspects of the problem dynamics that should be respected by an optimal policy. We test this hypothesis by:
\looseness -1
\begin {itemize}

  \item Defining a symbolic \partitioning\ method and establishing its basic theoretical properties. This method, SymPar, is adaptive to the problem semantics, general (i.e., not developed for a specific problem), and automatic (given a symbolically executable environment program).

  \item Implementing the method on top of the Symbolic PathFinder, an established symbolic executor for Java programs (JVM programs)\,\cite{pasareanu13ase}

  \item Evaluating SymPar empirically against other offline and online \partitioning\ approaches, and against deep reinforcement learning methods. The experiments show that symbolic \partitioning\ can allow the agent to learn better policies than with the baselines.

\end {itemize}

\noindent
To the best of our knowledge, this is the first time that  symbolic execution has been used to breath semantic knowledge into an otherwise statistical reinforcement learning process. We see it as an interesting case of a transfer of concepts from software engineering and formal methods to machine learning. It does break with the tradition of reinforcement learning to treat environments as black boxes. It is however consistent with common practice of using reinforcement learning for software defined problems and with pre-training robotic agents in simulators, as software problems and simulators are amenable to symbolic execution.
\looseness -1

The paper proceeds as follows.  \Cref{sec:relatedwork} reviews the relevant state of the art. \Cref{sec:preliminaries} recalls the required preliminaries and definitions. Our state space abstraction method is detailed in \cref{sec:partitioning}. In \cref{sec:Simulation Results} we present the evaluation design, and then discuss the experiment results (\cref{sec:results}). We discuss the limitations of our method in \cref{sec:Discussion}. Finally, \cref{sec:Conclusion} concludes the paper and presents future work. \looseness-1


\section{Related Work}
\label{sec:relatedwork}

We study \partitioning, or a discrete abstraction, of the state space in reinforcement learning by mapping from a continuous state space to a discrete one or by aggregating discrete states.  To the best of our knowledge, the earliest use of \partitioning, was the BOXES system\,\cite{michie1968boxes}. The Parti-game algorithm\,\cite{moore1991variable} automatically partitions state spaces but applies only to tasks with known goal regions and requires a greedy local controller. While tile coding is a classic method for \partitioning\,\cite{albus1981brains}, it often demands extensive engineering effort to avoid misleading the agent with suboptimal \partitions. Lanzi et al.\,\cite{lanzi2006classifier} extended learning classifier systems to use tile coding. Techniques such as vector quantization\,\cite {lee2004adaptive,wei2018q,mavridis2021vector,nicol2012states} and decision trees\,\cite{uther1998tree,Whiteson2010,seipp2018counterexample} lack adaptability to the properties of the state space and may overlook non-linear dependencies among state components. Techniques that gradually refine a coarse \partition\ during the learning process\,\cite{peter2019partitioning,wei2018q,akrour2018regularizing,mavridis2021vector,pmlr-v216-dadvar23a} are time-intensive, and require generating numerous \parts\ to achieve better approximations near the boundaries of nonlinear functions. Unlike other methods, SymPar incurs no direct learning cost (it is offline), requires no engineering effort (it is automated), and is not problem specific in contrast to some of the existing techniques (it is general). It produces a \partition\ that effectively captures non-linear dependencies as well as narrow \parts, without incurring additional costs or increasing the number of \parts\ at the boundaries.

The concept of bisimulation metrics \cite{ferns2004metrics,ferns2011bisimulation} defines two states as being behaviorally similar if they (1) yield comparable immediate rewards and (2) transition to states that are behaviorally aligned. Bisimulation metrics have been employed to reduce the dimensionality of state spaces through the aggregation of states. However, they have not been extensively explored due to their high computational costs. Moreover, note that bisimulation-minimization-based state-space-abstraction is too fine-grained for the problem at hand. It requires that any states lumped together exhibit the same behavior.  This is an unnecessary constraint from the reinforcement learning perspective, which takes no preference over behaviors provided that they lead to the same long-term reward.  As long as the same long-term reward estimate is expected for the same (best) local action in two states, it is theoretically sufficient for the two states to be lumped together.  For this reason it is worth exploring weaker principles than bisimulation metrics for reducing dimensionality.
\looseness -1



\section {Background}%
\label{sec:preliminaries}

\begin{figure}[t]
	\centering
	\begin{minipage}{0.48\textwidth}
		\centering
		\includegraphics [
		width = 0.7\linewidth,
		clip,
		trim = 3mm 2mm 2mm 1mm
		] {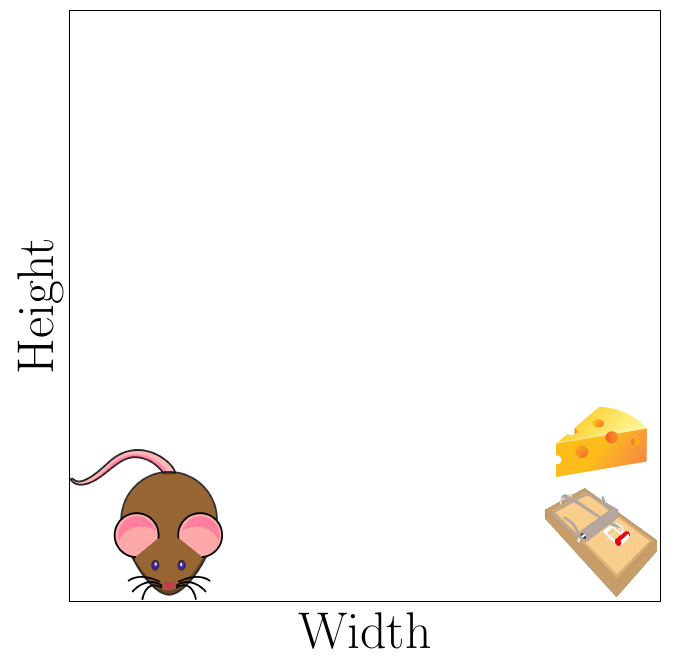}
		 \caption{Navigation environment. A mouse agent in a continuous rectangular board needs to find the cheese, while not stepping on the trap.}%
		\label{fig:navigationEnv}
	\end{minipage}
	\hfill
	\begin{minipage}{0.48\textwidth}
		\centering
		\includegraphics [
		width = \linewidth,
		clip,
		] {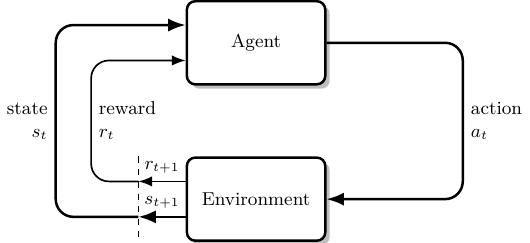}
		\caption{Reinforcement learning schematic.}%
		\label{fig:rl}
	\end{minipage}
\end{figure}


\paragraph {Reinforcement Learning} (see \cref{fig:rl}).
A Partially Observable Markov Decision Process is a tuple \( \mathcal M = (\EnvState, \IniState, \linebreak[3] \Action, \StateSet, \linebreak[2] \observe, \step\!, \reward, \isFinal)\), where \EnvState\ is a set of states, \( \IniState \in \PDF \EnvState \) is a probability density function for initial states, \Action is a finite set of actions, \StateSet is a finite set of observable states, \( \observe \in \EnvState \to \StateSet \) is a total observation function, \( \step \in \EnvState \times \Action \to \PDF \EnvState \) is the transition probability function, \( \reward \in \EnvState \times \Action \to \RR \) is the reward function, and \( \isFinal \in \StateSet \to \{ 0, 1 \} \) is a predicate defining final states. The task is to find a policy \(\pi: \StateSet \to \text{Dist} (\Action)\) that maximizes the expected accumulated reward \cite {sutton.barto:2018}, where Dist is the probability
mass
function over $\Action$.
\looseness -1

\begin {example*}\label{example:nav}
  A mouse sits in the bottom-left corner of a room with dimensions $W\!\times\! H$. A mousetrap is placed in the bottom-right corner, and a piece of cheese next to it (\cref{fig:navigationEnv}). The mouse moves with a fixed velocity in four directions: up, down, left, right. Its goal is to find the cheese but avoid the trap. The states \EnvState\ are ordered pairs representing the mouse's position in the room. The set of initial states \IniState is fixed to \( (1, 1) \), a Dirac distribution. We define the actions as the set of all possible movements for the mouse: \Action \( = \{(d, v): d \in \mathcal{D}, v \in \mathcal{V} \} \), where \( \mathcal{D}=\{U, D, R, L\} \) and \looseness -1 \( \mathcal{V} = \{r_1, r_2, \dots, r_n\ |\ r_i \in \mathbb{R}^+\} \). \StateSet can be any \partitioning\ of the room space and \observe is the map from the real position of the mouse to the \parti\ containing it. Our goal is to find the \partition, i.e., \StateSet and \observe. The reward function \reward is zero when mouse finds the cheese, \(-1000\) when the mouse moves into the trap, and \( -1 \) otherwise. For simplicity, we let the environment be deterministic, so \step is a deterministic movement of the mouse from a position by a given action to a new position. The final state predicate \isFinal holds for the cheese and trap positions and not otherwise.%
  \label{ex:nav}
  \looseness -1
\end {example*}

\paragraph{\Partitioning.}
\Partitioning\ is ``the process of mapping a representation of a problem onto a new representation'' \cite{giunchiglia1992theory}.
  A \emph{partition} over a set $\EnvState$ of states is a family of
  sets $p_1, \ldots, p_n\subseteq \EnvState$ such that
  $p_1\cup\ldots\cup p_n=\EnvState$
  and $p_i\cap p_j=\emptyset$ for $1\leq i<j\leq n$.
  The sets in a \partition\ are called \parts.  The set of all partitionings is partially ordered: we talk about coarseness (granularity) of partitions. A \partition\
  $\mathcal{P}'$ is \emph{coarser} than $\mathcal{P}$ (and
  $\mathcal{P}$ is \emph{finer} than $\mathcal{P}'$) if
  $\forall p\in\mathcal{P}.~\exists p'\in\mathcal{P}'.~ p \subseteq
  p'\ $. Recall that the space of partitions is isomorphic to the space of equivalence relations over a set.
  \looseness -1

\paragraph {Symbolic Execution} is a program analysis technique that systematically explores program behaviors by solving symbolic constraints obtained from conjoining the program's branch conditions\,\cite {king1976symbolic}. Symbolic execution extends normal execution by running the basic operators of a language using symbolic inputs (variables) and producing symbolic formulas as output. A symbolic execution of a program produces a set of \emph{path conditions}---logical expressions that encode conditions on the input symbols to follow a particular path in the program.

For a program over input arguments \( I = \{v_1, v_2, \dots, v_k\} \), a path condition \( \phi \in \PC (I^\prime)\) is a quantifier-free logical formula defined on $I^\prime = \{\vartheta_1, \vartheta_2, \dots, \vartheta_k\}$, where each symbolic variable $\vartheta_i$ represents an initial value for
$v_i$. \looseness -1

We briefly outline a definition of symbolic execution for a minimal
language (for more details, see, e.g., \cite {de2021symbolic}). Let
$V$ be a set of program variables, \Ops a set of arithmetic
operations, $x \in V$, $n \in \mathbb{R}$, and $op\in \Ops$. We
consider programs generated by the following grammar:
$$\begin{aligned}
    &e ::= x\ |\ n\ |\ op(e_1, \dots, e_n)
    \\
    &b ::= \True\ |\ \False\ |\ b_1 \text{ AND } b_2\ |\ b_1 \text{ OR } b_2\ |\  \neg b\ |\ b_1 \leq b_2\ |\ e_1 < e_2 ~|\ e_1 == e_2
    \\
    &s ::= x=e\ |\ x \sim \RND\ |\ s_1 ; s_2\ |\ \text{if } b:\ s_1\ \text{else}:\ s_2\ |\ \text{while } b:\ s\ |\ \text{skip}
\end{aligned}$$
A symbolic store, denoted by \( \sigma \) maps input program variables \( I \subseteq V \) to expressions, generated by productions \( e \) above. An update to a symbolic store is denoted
\(\sigma[x = e]\). It  replaces the entry for variable \(x\) with the expression \(e\). An expression can be interpreted in a symbolic store by applying (substituting) its mapping to the expression syntax (written \(e \sigma\)).

\begin{figure}[t]
    \begin{align*}
      \text{S-ASSIGN} & & (x=e, \sigma, k, \phi) &  ~~ \rightarrow ~~ (\text{skip}, \sigma[x\mapsto \sigma e], k, \phi)\\
      \text{S-IF-T} &  &(\text{if }b:\ s_1\ \text{else}: \ s_2, \sigma, k, \phi)  &  ~~ \rightarrow ~~  (s_1, \sigma, k, \phi \wedge\ \sigma b)\\
      \text{S-IF-F} &  &(\text{if }b:\ s_1\ \text{else}: \ s_2, \sigma, k, \phi)  &  ~~ \rightarrow ~~  (s_2, \sigma, k, \phi \wedge\ \sigma \neg b)\\
      \text{S-WHILE-T} & & (\text{while }b:\ s, \sigma, k, \phi)  &  ~~ \rightarrow ~~  (s\ ;\ \text{while }b:\ s, \sigma, k, \phi \wedge\ \sigma b)\\
      \text{S-WHILE-F} &  &(\text{while }b:\ s, \sigma, k, \phi)  &  ~~ \rightarrow ~~  (\text{skip}\ ;\ \sigma, k, \phi \wedge\ \sigma \neg b)\\
      \text{S-SMLP} &  &(x \sim \RND, \sigma, k, \phi) &  ~~ \rightarrow ~~ (\text{skip}, \sigma[x\mapsto y_k], k+1, \phi)
        \label{eq:semantics}
\end{align*}
\vspace{-4ex}
\caption {Symbolic execution rules for an idealized probabilistic language. Each judgement is a quadruple: the program, the symbolic store (\(\sigma\)), the sample index (\(k\)), the current path condition (\(\phi\)).}
\label{fig:semantics}
\vspace{-1ex}
\end{figure}

\Cref{fig:semantics} gives the symbolic execution rules for the above
language, in terms of traces (it computes a path condition \(\phi\)
for a terminating trace). In the reduction rules, \( \phi \)
represents the path condition and $k$ denotes the sampling index. The
first rule defines the symbolic assignment. An assignment does not
change the path conditions, but updates the symbolic store \(\sigma\).
When encountering conditional statements, the symbolic executor splits
into two branches. For the true case (rule S-IF-T) the path condition
is extended with the head condition of the branch, for the false case
(S-IF-F), the path condition is extended with the negation of the
branch condition. Similarly, for a \emph{while} loop two branches are
generated, with an analogous effect on path conditions. The last rule
executes the randomized sampling statement. It simply allocates a new
symbolic variable \(y_k\) for the unknown result of sampling, and
advances the sampling index\,\cite {erik2023}. \Cref{fig:navigationPC}
shows the path conditions obtained by applying similar rules to above
for the code to the left (\cref{fig:navigationProgram}). The first
path condition $PC^{(U,1)}$ corresponds to the branch where condition
\texttt{d==1} is true in the program.  \looseness -1

The above rules can be used to prove basic properties of symbolic
execution. For example, since branch conditions are always introduced
in dual rules,
the path conditions of a program are mutually exclusive \cite
{de2021symbolic}.

Practical symbolic executors have been realized for full scale
programming languages.  Although we defined symbolic execution at the
level of syntax, the two most popular symbolic executors operate on
compiled bytecode\,\cite{pasareanu13ase,cadar2008klee}. In presence of
loops and recursion, symbolic execution does not terminate. To halt
symbolic execution, we can set a predefined timeout in terms of an
iteration limit or a program statement limit. This produces an
approximation of the set of path conditions.  \looseness -1



\looseness -1







\section{\Partitioning\ Using Symbolic Execution}%
\label{sec:partitioning}

We present the idea of symbolic \partitioning\ using a single agent with the environment modeled as a computer program. The program (\Env) is implementing a single step-transition (\step) in the environment with the corresponding reward (\reward).  We use symbolic execution to analyze the environment program \Env, then \partition\ the state space using the obtained path conditions. The \partition\ serves as the observation function \observe. The entire process is automatic and generic---we can follow the same procedure for all problems.

\begin{figure}[t]
  \begin{center}
  \begin{minipage}{0.4 \linewidth}
    \input{NavProgram.tex}


    \caption{The environment program (\step, \reward) for the navigation problem (\cref{fig:navigationEnv}).}%
    \label{fig:navigationProgram}
  \end{minipage}
  \hspace{3mm}
  \begin{minipage}{0.5 \linewidth}
    
     \setlength {\tabcolsep} {2pt}
    \begin{tabular} {
      >{\tiny}rl
      >{\footnotesize}l
    }
         & \(\PC^{(U, 1)}\) \\
         \textcolor{gray}{5,\,19,\,20} &                  \(y\!<\!10 \land x\!=\!10\,\land y\!+\!1\!=\!2\)\\
         \textcolor{gray}{5,\,19,\,!20} &                  \(y\!<\!10 \land x\!=\!10 \land y\!+\!1\!\neq\!2\)\\
         \textcolor{gray}{5,\,!19} &                  \(y\!<\!10 \land x\!\neq\!10\)\\
         \textcolor{gray}{!5,\,19} &                  \(y\!\geq\!10 \land x\!=\!10\)\\
         \textcolor{gray}{!5,\,!19} &                  \(y\!\geq\!10 \land x\!\neq\!10\)
        \\[2mm]

         & \(\PC^{(D, 1)}\) \\
         \textcolor{gray}{8,\,19,\,20} &                  \(y\!>\!1 \land x\!=\!10 \land y\!-\!1\!=\!2\)\\
         \textcolor{gray}{8,\,19,\,!20,\,22} &                  \(y\!>\!1 \land x\!=\!10 \land y\!-\!1\!\neq\!2 \land y\!-\!1\!=\!1\)\\
         \textcolor{gray}{8,\,19,\,!20,\,!22} &                  \(y\!>\!1 \land x\!=\!10 \land y\!-\!1\!\neq\!2 \land y\!-\!1\!\neq\!1\)\\
         \textcolor{gray}{8,\,!19} &                  \(y\!>\!1 \land x\!\neq\!10\)\\
         \textcolor{gray}{!8,\,19,\,!20,\,22} &                  \(y\!\leq\!1 \land x\!=\!10 \land y\!=\!1\)\\
         \textcolor{gray}{!8,\,!19} &                  \(y\!\leq\!1 \land x\!\neq\!10\)
        \\[2mm]

         & \(\PC^{(L, 1)}\)\\
         \textcolor{gray}{11,\,!19} &                  \(x\!>\!1 \land x\!-\!1\!\neq\!10\)\\
         \textcolor{gray}{!11,\,!19} &                  \(x\!\leq\!1 \land x\!\neq\!10\)
        \\[2mm]

         & \(\PC^{(R, 1)}\) \\
         \textcolor{gray}{14,\,19,\,20} &                  \(x\!<\!10 \land x\!+\!1\!=\!10 \land y\!=\!2\)\\
         \textcolor{gray}{14,\,19,\,!20,\,22} &                  \(x\!<\!10 \land x\!+\!1\!=\!10 \land y\!\neq\!2 \land y\!=\!1\)\\
         \textcolor{gray}{14,\, 19,\,!20,\,!22} &                  \(x\!<\!10 \land x\!+\!1\!=\!10 \land y\!\neq\!2 \land y\!\neq\!1\)\\
         \textcolor{gray}{14,\,!19} &                  \(x\!<\!10 \land x\!+\!1\!\neq\!10\)\\
         \textcolor{gray}{!14,\,19,\,20} &                  \(x\!\geq\!10 \land x\!=\!10 \land y\!=\!2\)\\
         \textcolor{gray}{!14,\,19,\,!20,\,22} &                  \(x\!\geq\!10 \land x\!=\!10 \land y\!\neq\!2 \land y\!=\!1\)\\
         \textcolor{gray}{!14,\,19,\,!20,\,!22} &                  \(x\!\geq\!10 \land x\!=\!10 \land y\!\neq\!1 \land y\!\neq\!1\)\\[-0.8mm]\smash\strut
      \end{tabular}%
      \caption{Path conditions collected by symbolic execution. The numbers (to the left) refer to line numbers in the program of \cref{fig:navigationProgram}.}%
      \label{fig:navigationPC}
    \end{minipage}
  \end{center}
  \vspace{-6mm}
\end{figure}

\begin{example*}
  \Cref {fig:navigationProgram} shows the environment program for the \( 10 \times 10 \) navigation problem (\cref{example:nav}). For simplicity, we assume the agent can move one unit in each direction, so \( \mathcal V = \{ 1 \} \) and \( \Action= \{U, D, R, L \} \times \mathcal V \). The path conditions in \cref{fig:navigationPC} are obtained by symbolically executing the step and reward functions using symbolic inputs \( x \) and \( y \) and a concrete input from \Action. Using path conditions in \partitioning\ requires a translation from the symbolic executor syntax into the programming language used to implement the \partitioning\ process, as the executor will generate abstract value names.
  \looseness -1
\end{example*}

\noindent
A good \partition\ maintains the Markov property, so that the same action is optimal for all unobservable states abstracted by the same \parti. Unfortunately, this means that a good \partition\ can be selected only once we know a good policy---after learning. To overcome this, SymPar heuristically bundles states into the same \parti\ if they induce the same execution path in the environment program. We use an off-the-shelf symbolic executor to extract all possible path conditions from \Env, by using \EnvState\ as symbolic input and actions from \Action as concrete input.  The result is a set \( \PC \) of path conditions for each concrete action: \( \PC = \{\PC^{a_0}, \PC ^{a_1}, \dots, \PC ^{a_m} \} \), where \(\PC^a = \{\PC^a_0, \PC^a_1, \dots, \PC^a_{k_a}\}\). The set \( \PC^a \) contains the path conditions computed for action \( a \), and \( k_a \) is the number of all path conditions obtained by running \Env\ symbolically, for a concrete action \( a \).
\looseness -1

Running the environment program for any concrete state satisfying a condition \(\PC^a_i\) with action \(a\) will execute the same program path.  However, the \partitioning\ for reinforcement learning needs to be action independent (the same for all actions).  So the obtained path conditions cannot be used directly for the \partitioning.
Consider \( \PC_i^{a_1} \in \PC^{a_1} \) and \( \PC_j^{a_2} \in \PC^{a_2} \), arbitrary path conditions for some actions \(a_1\), \(a_2\). To make sure that the same program path will be taken from a concrete state for both actions, we need to create a \parti\ that corresponds to the intersection of both path conditions: \( \PC_i ^{a_1} \land \PC_j ^{a_2} \).  In general, each set in $\PC$ defines \partitions\ of the state space for different actions.  To make them compatible, we need to compute the unique coarsest \partition\ finer than those induced by the path conditions for any action, which is a standard operation in order theory \cite {davey1990introduction}. In this case, this amounts to computing all intersections of \partitions\ for all actions, and removing the empty intersections using an SMT check.
\looseness -1

\begin {figure} [t!]
    \centering
\scalebox{0.89}{\begin{tikzpicture}
  \node (stepFunc) [cylinder,
    draw=black,
    shape border rotate=90,
    minimum height=2cm,
    minimum width = 16mm,
    text width=14mm,
    inner sep = 1mm,
    align=center,
    shape aspect=.25,] {Simulator Code \\(\step, \reward)};

  \node (arrow1) [single arrow,
    draw=black,
    very thick,
    minimum width = 20pt,
    single arrow head extend=5pt,
    minimum height=8mm,
    right = 1mm of stepFunc,] {};

  \node (SE) [rectangle,
    rounded corners=5,
    draw=black,
    minimum height=2cm,
    minimum width=16mm,
    inner sep = 1mm,
    text width = 14mm,
    align=center,
    right = 1mm of arrow1,] {Symbolic Executor};

  \node (arrow2) [single arrow,
    draw=black,
    very thick,
    minimum width = 20pt,
    single arrow head extend=5pt,
    minimum height=8mm,
    rotate = -90,
    xshift = -5mm,
    yshift = -2mm,
    above= 2mm of SE,] {};

  \node (actions) [rectangle,
    above = 2mm of arrow2,
    draw=black,
    xshift = -2mm,
    yshift = 3mm,
    ] {$a_1, a_2, \dots, a_m$};

  \node (arrow3) [single arrow,
    draw=black,
    very thick,
    minimum width = 20pt,
    single arrow head extend=5pt,
    minimum height=8mm,
    right = 1mm of SE,] {};

  \node (border1) [rectangle,
    draw,
    dashed,
    inner sep = 1mm,
    right= 1mm of arrow3]{\begin{tikzpicture}[solid]
        \node (pc1I) [rectangle,
          draw=black,] {\includegraphics[scale = 0.05]{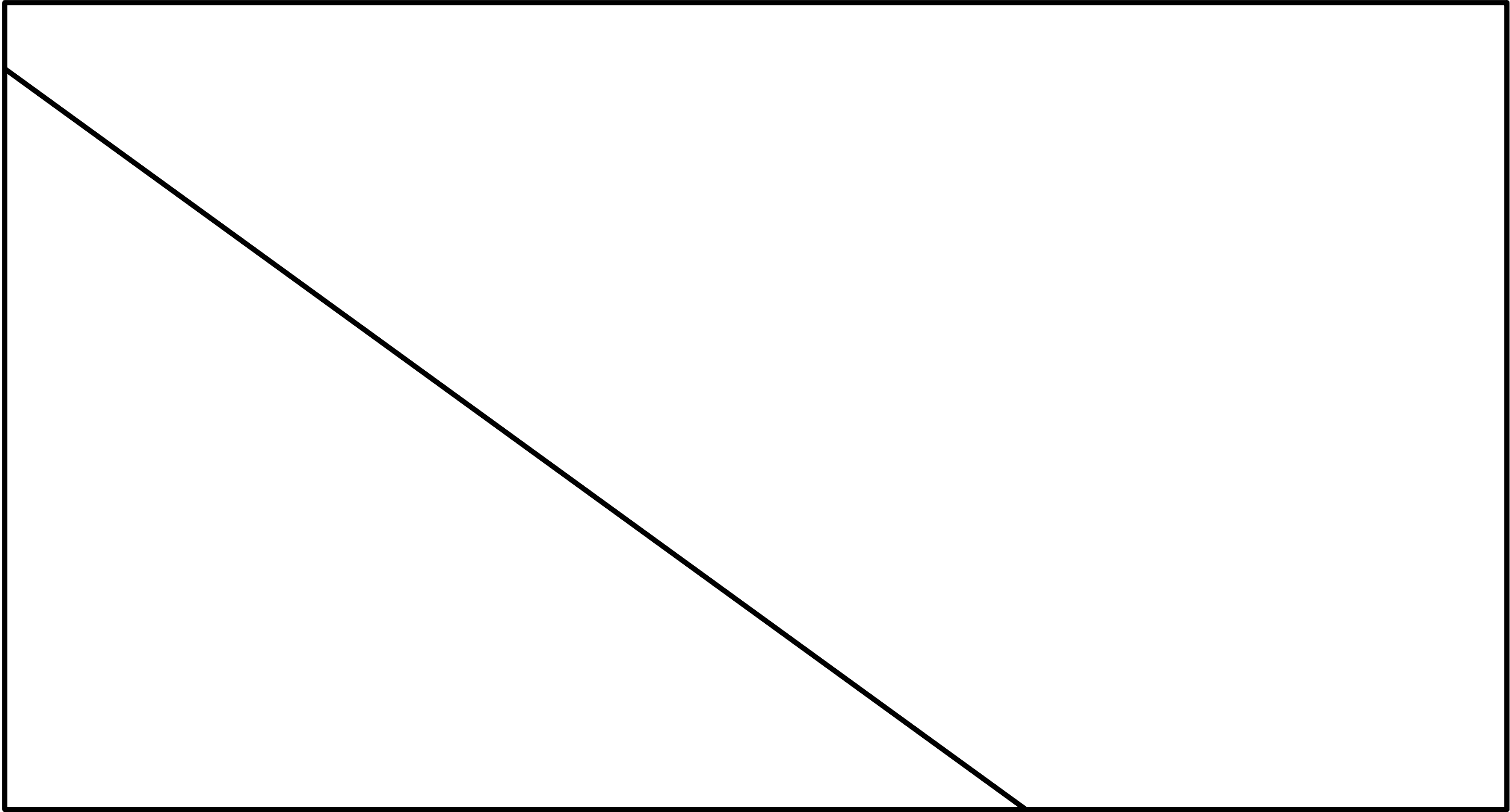}};
        \node (pc1) [rectangle,
          left = 0mm of pc1I,] {$PC^{a_1}$};

        \node (pc2I) [rectangle,
          draw=black,
          below= 2mm of pc1I] {\includegraphics[scale = 0.05]{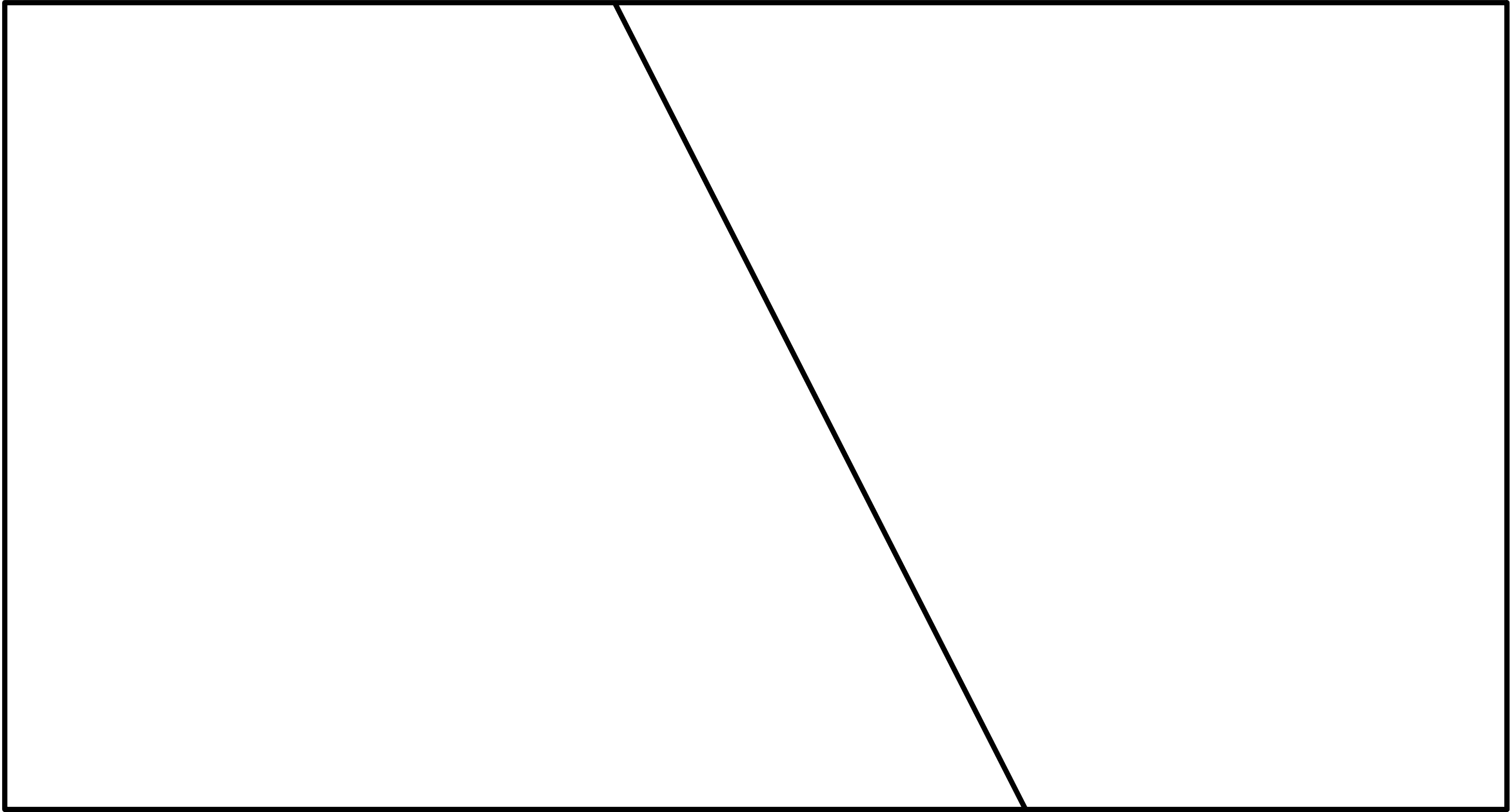}};
        \node (pc2) [rectangle,
          left = 0mm of pc2I,] {$PC^{a_2}$};

        \node (pc3I) [rectangle,
          draw=black,
          below = 2mm of pc2I,] {\includegraphics[scale = 0.05]{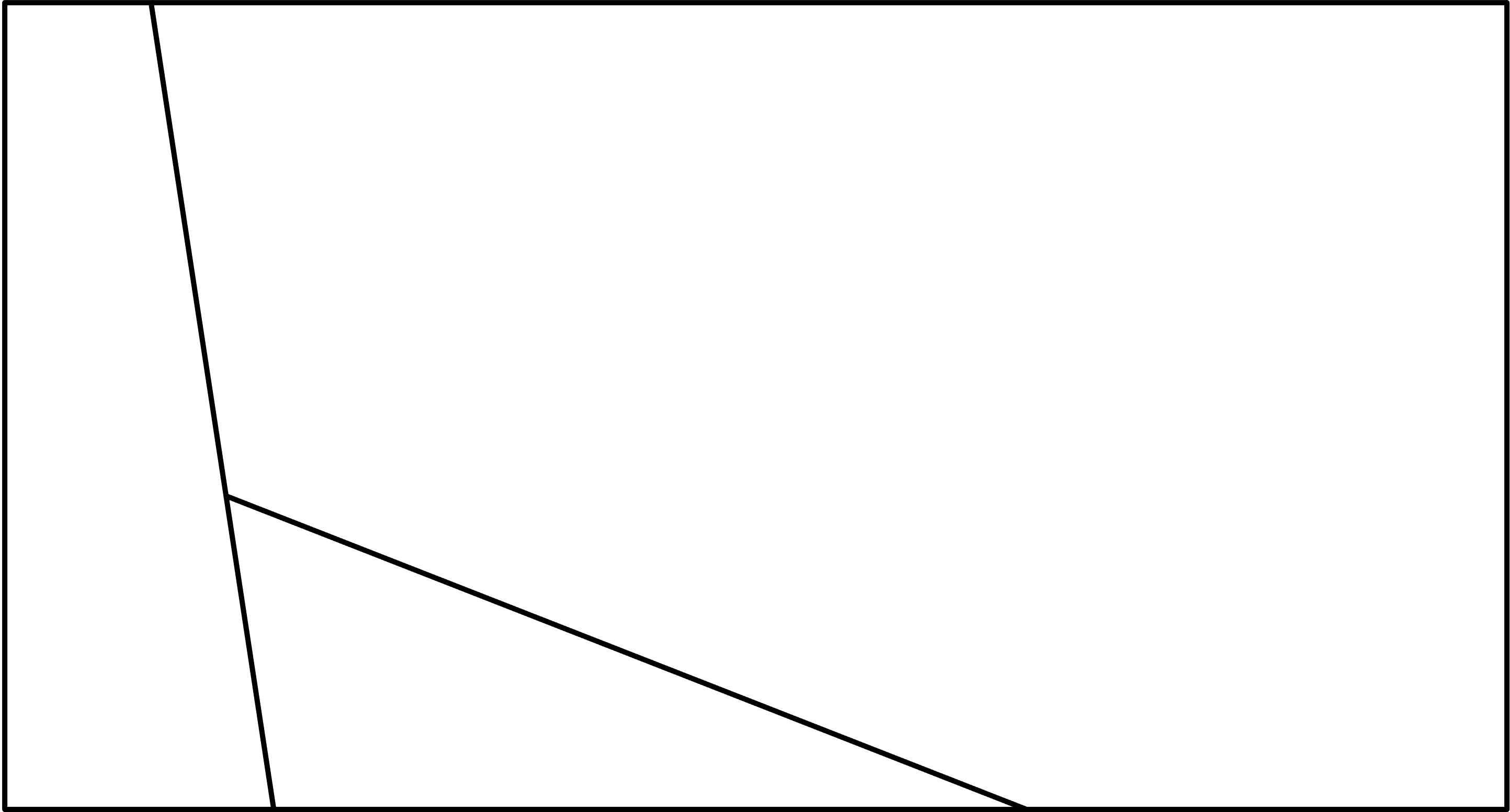}};
        \node (pc3) [rectangle,
          left = 0mm of pc3I,] {$PC^{a_3}$};

        \node (pc4I) [rectangle,
          draw=black,
          below= 2mm of pc3I] {\includegraphics[scale = 0.05]{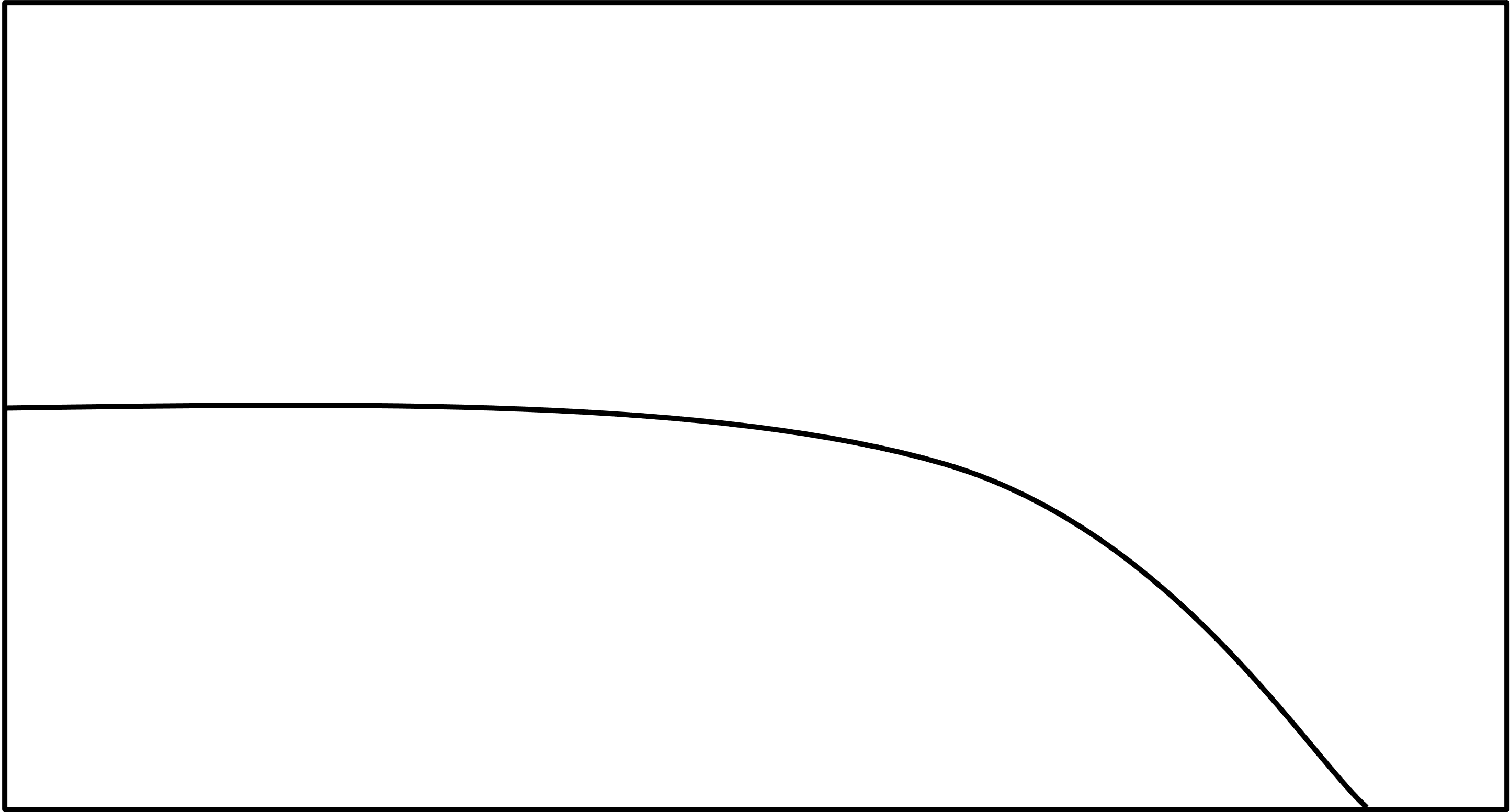}};
        \node (pc4) [rectangle,
          left = 0mm of pc4I,] {$PC^{a_4}$};

        \node (pc5I) [rectangle,
          draw=black,
          below = 2mm of pc4I,] {\includegraphics[scale = 0.05]{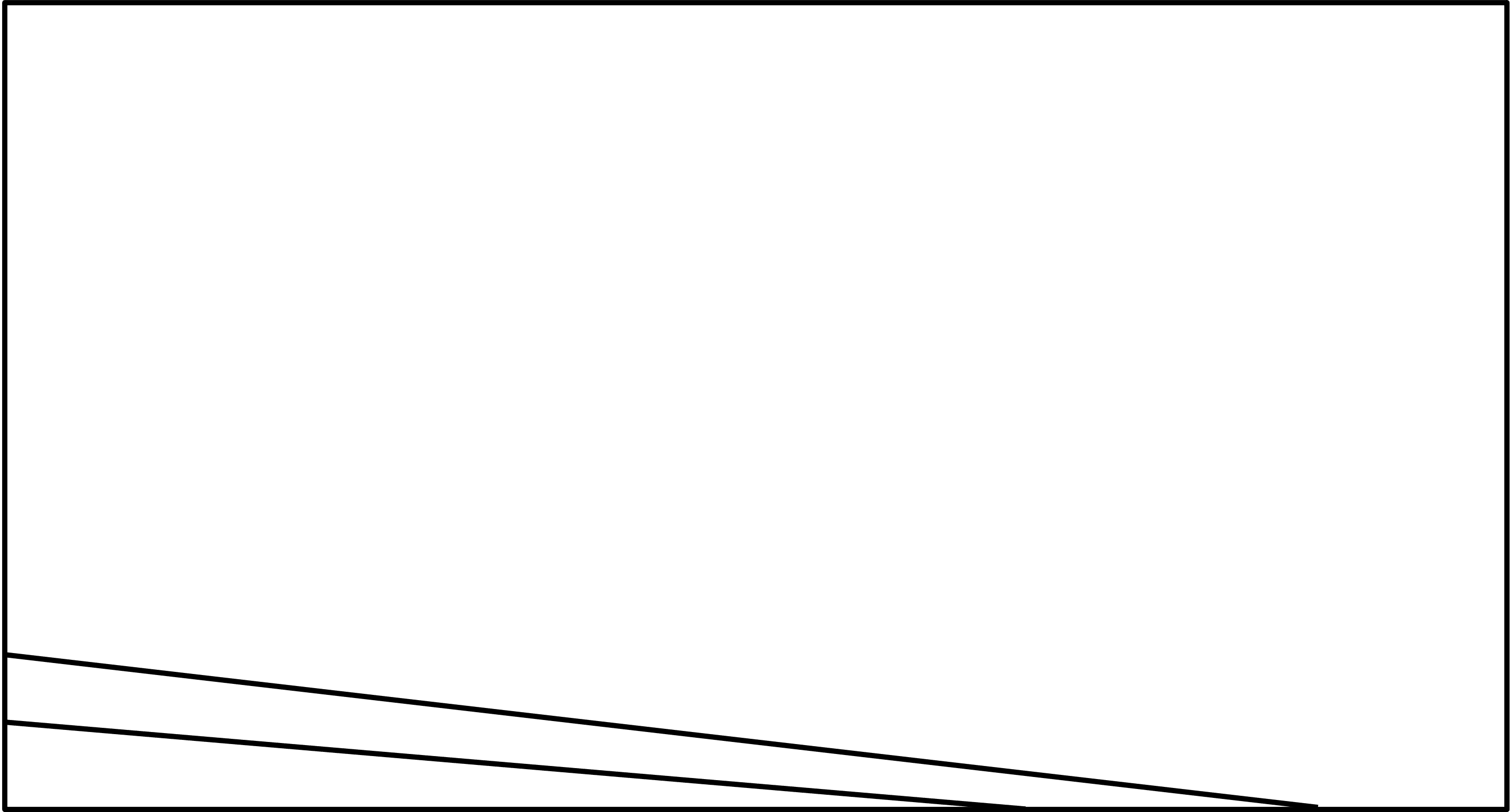}};
        \node (pc5) [rectangle,
          left = 0mm of pc5I,] {$PC^{a_5}$};

    \end{tikzpicture}};

  \node (arrow4) [single arrow,
  draw=black,
  very thick,
  minimum width = 20pt,
  single arrow head extend=5pt,
  minimum height=8mm,
  right = 1mm of border1,] {};

  \node (res) [rectangle,
  align = left,
  right = 0mm of arrow4] {\includegraphics[scale = 0.1]{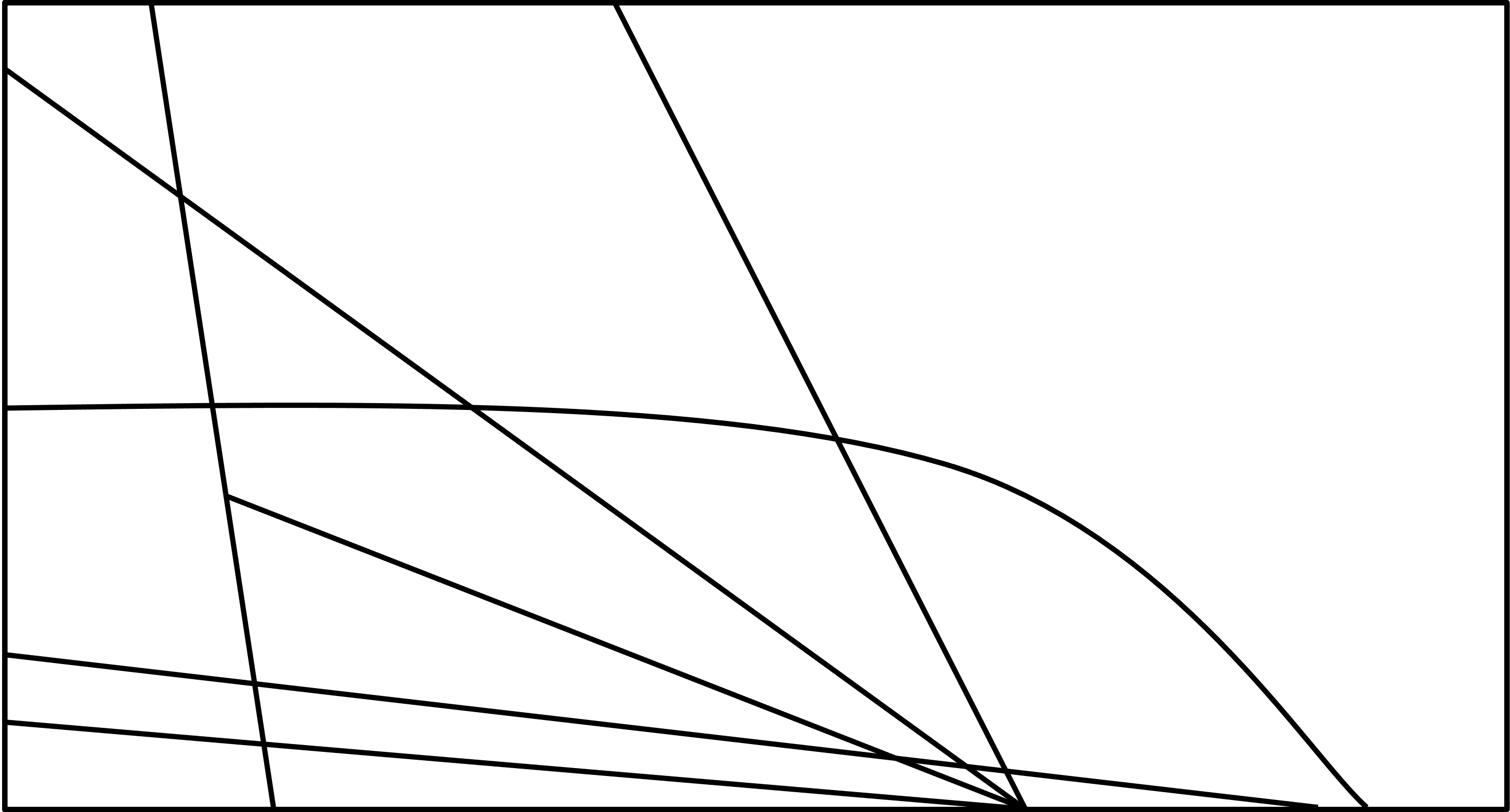}};

\end{tikzpicture}}

\caption{Overview of SymPar.}%
\label{diagram1}




\end {figure}

\begin {figure} [t!]
\centering
    \includegraphics[scale = 0.12]{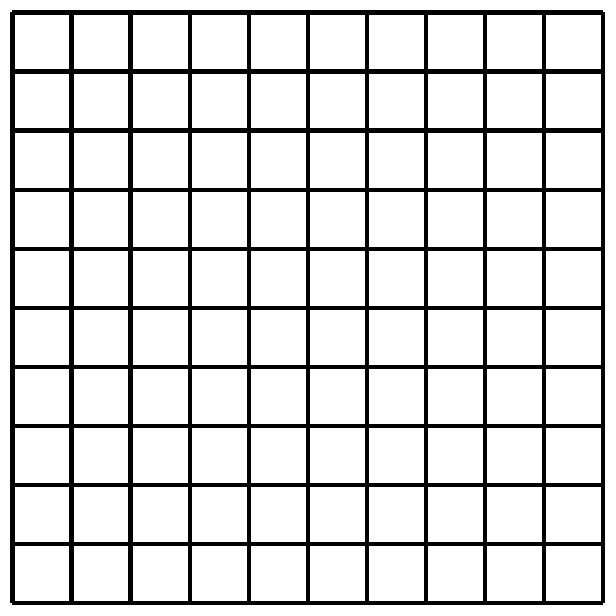}
    \includegraphics[scale = 0.068]{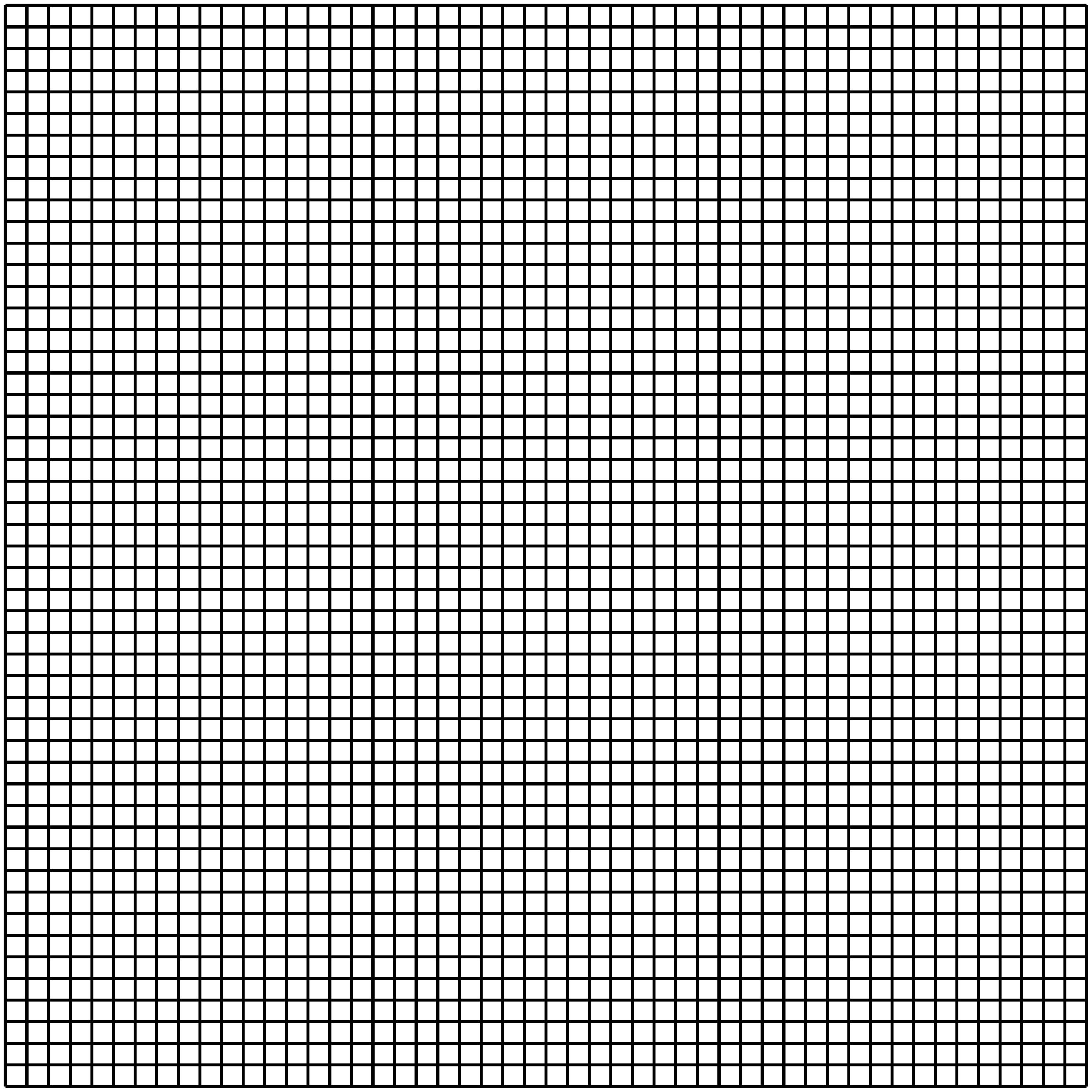}
    \includegraphics[scale = 0.12]{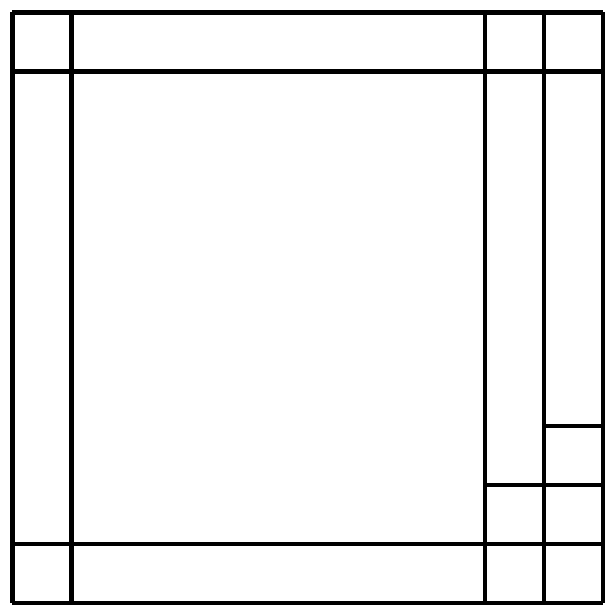}
     \includegraphics[scale = 0.068]{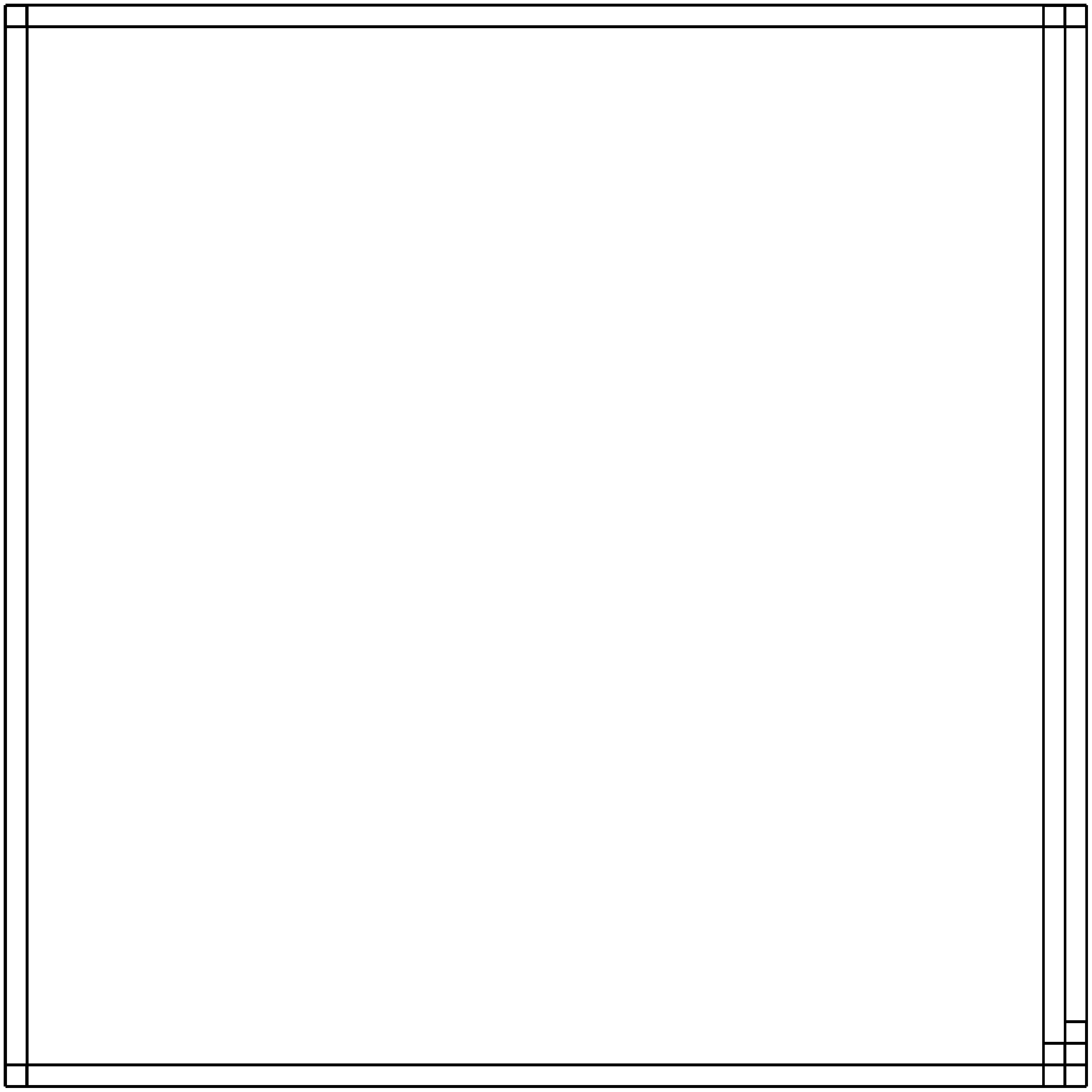}


  \caption {Using tile coding (left) and SymPar (right) for \( 10 \! \times \! 10 \) and \( 50 \! \times \! 50 \) navigation}%
  \label{fig:navigationTiling}

\end {figure}

The process of symbolic state space \partitioning\  is summarized in \cref{diagram1,alg:sympar}.  SymPar executes the environment program symbolically. For each action, a set of path conditions is collected. In the figure, \( |\Action| = 5 \) and, accordingly, five sets of path conditions are collected (shown as rectangles). Each rectangle is divided into a group of regions, each of which maps to a path condition. Thus, the rectangles illustrate the state space that is discretized by the path conditions. Note that the border of each region can be a unique path condition (an expression with equality relation) or a \parti\ of neighbour regions (an expression with inequality relation). The final \partition\ is shown as another rectangle that contains the overlap between the regions from the previous step.\looseness -1

\begin {example*}
\Cref {fig:navigationTiling} (left) shows the \partitioning\ of the Navigation problem using tile coding \cite {albus1981brains} for two room sizes. Numerous cells share the same policy, prompting the question of why they should be divided. SymPar achieves a much coarser \partition\ than the initial tiling, by discovering that for many tiles the dynamics is the same (right).
\end {example*}

\noindent
We handle stochasticity of the environment by allowing environment programs to be probabilistic and then following rule S-SMLP in symbolic execution (\cref{fig:semantics}). We introduce a new symbolic variable whenever a random variable is sampled in the program\,\cite{erik2023,dexter1979}.  Consequently, our path conditions also contain these sampling variables.  To make the process more reliable, one can generate constraints, limiting them to the support of the distribution. For example, for sampling from a uniform distribution $U[\alpha, \beta]$, the sampling variable $n_v$ is subject to two constraints: $n_v\!\geq\!\alpha$ and $n_v\!\leq\!\beta$.  In order to be able to compute the \partition\ over state variables only, as above, we existentially quantify the sampling variables out. This may introduce overlaps between the conditions, so we compute their intersection at this stage before proceeding (see lines 6-9 in \cref{alg:sympar}).
\looseness -1


Since the entire setup uses logical representations and an SMT solver, we exploit it further to generate witnesses for all \parts, even the smallest ones.  We use them to seed reinforcement learning episodes, ensuring that each \parti\ has been visited at least once. Consequently the agent is guaranteed to learn from all the paths in the environment program. This can be further improved by constraining with a reachability predicate (not used in our examples).
\looseness -1

\begin{algorithm}[tb]
    \caption{Partitioning with Symbolic Execution (SymPar)}%
    \label{alg:sympar}
    \textbf{Input}: \Env, \Action\\
    \textbf{Output}: $\mathcal{P}$ (a partitioning of \EnvState)
    \begin{algorithmic}[1] 
        \State $\PC\leftarrow\emptyset$
        \For {$a\in{}$\Action}
        \State $\PC^a, \Psi \leftarrow$ SymExec\ (\Env, symbolic \EnvState, concrete $a$) \hspace{2mm} //  $\Psi$ \emph{is the set of sampling variables}
        \State Add distribution support constraints for all variables \(\EnvState \in \Psi\) to \( \PC^a \)
        \State Existentially quantify all sampling variables in \( \PC^a \) \hspace{3mm} // \emph{may introduce overlaps of conditions}
        \State $\PC^{\prime a}\leftarrow\emptyset$
        \For {$p, q\in \PC^a$} \textbf{if} {$\textbf{SAT}\ (p \wedge q)$} \(PC^{\prime a} \gets PC^{\prime a} \cup \{p \land q\}\) \EndFor
        \State $\PC^{a} \leftarrow PC^{\prime a}$
        \EndFor
        \State $\mathcal{P} \leftarrow \PC^{\mathcal{A}[0]}$
        \For {$a\in\Action - \{\mathcal{A}[0]\}$}
        \State $\mathcal{P}^\prime\leftarrow\emptyset$
        \For {$p\in\mathcal{P}, q\in \PC^a$}
        \textbf{if} $\textbf{SAT}\ (p \wedge q)$ \textbf{then} $\mathcal{P}^\prime \gets \mathcal{P}^\prime \cup \{ p \land q \}$
        \EndFor
        \State $\mathcal{P} \leftarrow \mathcal{P}^\prime$
        \EndFor
       	\State $\mathcal{P} \leftarrow \mathcal{P} \cup \{\sim\bigvee_{p\in\mathcal{P}} p\}$
        \State \textbf{return} $\mathcal{P}$
    \end{algorithmic}
\end{algorithm}


\paragraph{Properties of SymPar.}%
\label{Sec:Analysis}

SymPar on the specifics of the environment implementation. Distinct
implementations of the simulated environment may result in different
\partitioning\ outcomes for a given problem. On the other hand, the outcome is
independent of the size of state space. Recall that in
\cref{fig:navigationTiling} (right) the number of \parts\ is the same for
the small and the large room.

A \partition\ is by definition total: every state in the input space is included in a \parti, ensuring the entire state space is fully covered.  As symbolic execution does not terminate for many interesting programs (programs with loops have infinitely many symbolic paths), one typically stops symbolic execution after a designated timeout.  This can leave a part of the state space unexplored. Hence, a partitioning obtained from path conditions generated by symbolic execution may not cover all the state space.  SymPar makes the obtained partition total by adding the complement of the union of the computed partitions, to cover for the unexplored paths (l.\,14 in \cref{alg:sympar}). Thus, the following property holds:
\looseness -1

\begin{theorem}
\label{th:total}
The set  \( \mathcal P \)obtained in \cref{alg:sympar} is a partition (i.e., it is total):
	\(\forall \EnvS \in \EnvState \; \exists!\, \mathcal{P}_0 \in\mathcal{P}\cdot \; \EnvS \in \mathcal{P}_0.\)
\end{theorem}

\noindent
The cost of SymPar amounts to exploring all paths in the program symbolically and then computing the coarsest \partition. The symbolic execution involves generating a number of paths exponential in the number of branch points in the program (and at each branch one needs to solve an SMT problem---which is in principle undecidable, but works well for many practical problems). A practical approach is to bound the depth of exploration of paths by symbolic executor for more complex programs. Computing the coarsest \partition\ requires solving $|\mathcal P|^{|\mathcal A|}$ number of SMT problems where $|\mathcal P|$ is the upper bound on the number of \parts\ (symbolic paths) and $|\mathcal A|$ is the number of actions. The other operations involved in this process such as computing and storing the path conditions in the required syntax are polynomial and efficient in practice.
\looseness -1

\begin {theorem}
\label{th:time}
Let \( \PC ^a \) be the set of path conditions produced by SymPar for
each of the actions \( a \in \Action \). The size of the final
\partition\ \Pcal returned by SymPar is bounded from below by each
\(|\PC ^a|\) and from above by \( \prod_{a\in\Action}|\PC^a| \).
\end {theorem}

\noindent
The theorem follows from the fact that \Pcal is finer than any of
the \( \PC ^a \)s and the algorithm for computing the coarsest
\partition\ finer
than
a set of \partitions\ can in the worst case
intersect each \parti\ in each set \( PC^a \) with all the
\parts\ in the \partitions\ of the other actions.

Note that SymPar is a heuristic and approximate method. To appreciate this, define the optimal \partition\ to be the unique \partition\ in which each \parti contains all states with the same action in the optimal policy (the optimal \partition\ is an inverse image of the optimal policy for all actions). The \partitions\ produced by SymPar are neither always coarser or always finer than the optimal one. This can be shown with simple counterexamples. For an environment with only one action, the optimal \partition\ has only one \parti\ as the optimal policy maps the same action for all states. But Sympar will generate more than one \parti\ (a finer \partition) if the simulation program contains branching. For problems without branching in the simulator such as cart pole problem, Sympar produces only one \parti. However, the optimal \partition\ contains more than one \parti\ as optimal actions for all states in the state space are not the same.  To understand the significance of this approximation in practice, we evaluate SymPar empirically against the existing methods.
\looseness -1

\section{Evaluation Setup}%
\label{sec:Simulation Results}

The \partitioning\ of the state space faces a trade-off: on one hand,
the granularity of the \partition\ should be fine enough to
distinguish crucial differences between states in the state space. On
the other hand, this granularity should be chosen to avoid a
combinatorial explosion, where the number of possible \parts\
becomes unmanageably large. Achieving this balance is essential for
efficient and effective learning. In this section, we explore this
trade-off and evaluate the performance of our implementation in SymPar
empirically by addressing the following research questions:
\looseness -1

\medskip

\noindent
\begin{tabularx} \linewidth {@{}lX@{}}

  \textbf{RQ1}
  & \emph{How much smaller are the SymPar \partitions\ compared to other methods, \strut\,and how do these smaller \partitions\ impact learning performance?}
  \\

  \textbf{RQ2} & \strut \emph{How does the granularity of the \partition\ affect the learning performance?} \\

  \textbf{RQ3} & \strut \emph {How does SymPar scale with increasing state space sizes?} \\

  \textbf{RQ4} & \strut \emph {How well does SymPar group together behaviorally similar states?}\\[0ex]

\end{tabularx}

\smallskip

\noindent
We compare SymPar to CAT-RL\,\cite{pmlr-v216-dadvar23a} (online \partitioning) and with tile coding techniques (offline \partitioning) for different examples\,\cite{sutton.barto:2018}.  Tile coding is a classic technique for \partitioning. It splits each feature of the state into rectangular tiles of the same size.  Although there are new problem specific versions, we opt for the classic version due to its generality.
\looseness -1


To answer \textbf{RQ1},
we measure (a) the \emph{size of \partition}, (b) the \emph{failure
  and success rates} and (c) the \emph{accumulated reward} during
learning.  Being offline, our approach is hard to compare with online
methods, since the different parameters may affect the results.
Therefore, we separate the comparison with offline and online
algorithms. For offline algorithms, we first find the number of
abstract states using SymPar and \partition\ the state space using tile
coding accordingly (i.e., the number of tiles is set to the smallest
square number greater than the number of \parts\ in
SymPar's \partition). Then, we use standard Q-learning for these \partitions, and
compare their performance.
%
For online algorithms, we compute the running time for SymPar and its
learning process, run CAT-RL for the same amount of time, and
compare their performance. Obviously, if the agent observes a failing
state, the episode stops. This decreases the running time.
Finally, we compare the accumulated reward for SymPar with
well-known algorithms DQN \cite{mnih2013playing}, A2C
\cite{mnih2016asynchronous}, PPO \cite{schulman2017proximal},
using the Stable-Baselines3
implementations\footnote{https://github.com/DLR-RM/stable-baselines3}
\cite{raffin2019stable}.  These comparisons are done for two
complementary cases: (1) randomly selected states and (2) states that
are less likely to be chosen by random selection. The latter are
identified by SymPar's \partition.
We sample states from different \parts\ obtained by SymPar and
evaluate the learning process by measuring the accumulated reward.
\looseness -1

To answer \textbf{RQ2}, we create different learning
problems with
various \partitioning\ granularities
by changing the search depth for the symbolic execution.  We then
compare the maximum accumulated reward of the learned policy to gain an understanding of the learning
performance for the given abstraction.\looseness -1

To answer \textbf{RQ3}, we compare the number of \parts\ when increasing the state space of problems.

To answer \textbf{RQ4}, we select five random \parts\ from the \partition\ obtained by SymPar, and five random concrete states from each \parti. Then, we feed the concrete states as initial states to RL, and compute the accumulated reward using the policy obtained from a trained model, assuming the training converged to the optimal policy. This way we can check how different the concrete states are with regard to performance.

\paragraph{Test Problems.} The \textbf{Navigation} problem
with a room (continuous) size of $10\!\times\!10$.  The \textbf{Simple Maze} is a discrete environment ($100\!\times\!100$) including blocks, goal and trap, in which a robot tries to find the shortest and safest route to the goal state \cite{sutton.barto:2018}.  \textbf{Braking Car} describes a car moving towards an obstacle with a given velocity and distance. The goal is to stop the car to avoid a crash with minimum braking pressure \cite{Varshosaz2023}.  The \textbf{Multi-Agent Navigation} environment ($10\!\times\!10$ grid) contains two agents attempting to find safe routes to a goal location. They must arrive to the goal position at the same time \cite{sharon2015conflict}.  The \textbf{Mountain Car} aims to learn how to obtain enough momentum to move up a steep slope\,\cite{mountaincar}.  The \textbf{Random Walk} in continuous space is an agent with noisy actions on an infinite line \cite{sutton.barto:2018}. The agent aims to avoid a hole and reach the goal region.
\textbf{Wumpus World} \cite{russell2010artificial} is a grid world (1: $64\!\times\!64$, 2: $16\!\times\!16$) in which the agent should avoid holes and find the gold.

\section{Results}
\label{sec:results}

\begin{table}[t]
	\centering
	
	\setlength {\tabcolsep} {2.5pt}
	\begin{tabular}
		{@{\hspace{0pt}}
			>{\footnotesize}p{8.5mm}@{\hspace{1pt}}
			>{\footnotesize}r@{\hspace{.3em}}
			>{\footnotesize}r
			>{\footnotesize}r
			>{\footnotesize}r
			>{\footnotesize}r@{\hspace{3pt}}|
			>{\footnotesize}r@{\hspace{.3em}}
			>{\footnotesize}r@{\hspace{.3em}}
			>{\footnotesize}r
			>{\footnotesize}r@{\hspace{.3em}}
			>{\footnotesize}r@{\hspace{3pt}}
			|
			>{\footnotesize}r@{\hspace{.3em}}
			>{\footnotesize}r@{\hspace{.3em}}
			>{\footnotesize}r
			>{\footnotesize}r@{\hspace{.3em}}
		}
		&\multicolumn{5}{p{35mm}|}{\centering\footnotesize\textbf{SymPar}}
		&\multicolumn{5}{p{35mm}}{\centering\footnotesize\textbf{Tile Coding}}&\multicolumn{4}{p{28mm}}{\centering\footnotesize\textbf{CAT-RL}}\\[-.6mm]
		&\multicolumn{1}{p{5.8mm}}{\centering\footnotesize\bm{{|\StateSet{}|}}}
		&\footnotesize\textbf{Succ}
		&\footnotesize\textbf{Fail}
		&\footnotesize\textbf{$\textbf{T}_\textbf{out}$}
		&\footnotesize\textbf{\llap Opt}
		&\footnotesize\bm{{|\StateSet{}|}}
		&\footnotesize\textbf{Succ}
		&\footnotesize\textbf{Fail}
		&\footnotesize\textbf{$\textbf{T}_\textbf{out}$}
		&\footnotesize\textbf{Opt}
		&\multicolumn{1}{p{5.4mm}}{\footnotesize\bm{{|\StateSet{}|}}}
		&\multicolumn{1}{p{7mm}}{\footnotesize\textbf{Succ}}
		&\multicolumn{1}{p{7mm}}{\footnotesize\textbf{Fail}}
		&\footnotesize\textbf{$\textbf{T}_\textbf{out}$}\\    &\scriptsize{(\#)}&\scriptsize{(\%)}&\scriptsize{(\%)}&\scriptsize{(\%)}&\scriptsize{(\%)}&\scriptsize{(\#)}&\scriptsize{(\%)}&\scriptsize{(\%)}&\scriptsize{(\%)}&\scriptsize{(\%)}&\scriptsize{(\#)}&\scriptsize{(\%)}&\scriptsize{(\%)}&\scriptsize{(\%)}
		\\
		\midrule
		\textbf{SM}  & 33 & 74.9 & <0.1 & 25.0 & 5.0 & $10^4$ & 6.0 & 7.1 & 86.9& 0.0 &154 & 63.0 & 5.0 &32.0\\
		\textbf{MAN} & 130 & 5.8 & 82.6 & 11.6 & 0.0 & $10^4$ & 0.0 & 99.6 & 0.4& 0.0  &620 & 0.0 & 74.7 &25.3\\
		\textbf{WW\,1} & 73 & 18.4 & 0.0 & 81.6 & 2.1 & $8^4$ & 9.6 & 0.0 & 90.4& 0.0  & 157 & 2.7 & 0.0 &97.3\\
		\textbf{WW\,2} & 52 & 37.3 & 22.9 & 39.8 & 4.2 & 64 & 19.1 & 33.2 & 47.7& 0.0  & 22 & 14.5 & 30.2 &55.3\\
		\midrule
		\textbf{Nav} & 51 & 13.2 & 4.8 & 82.0 & <0.1 & 64 & 0.0 & 0.0 &100.0& 0.0  & 100 & 1.7 & 1.5 & 96.8\\
		\textbf{BC} & 81 & 89.1 & 10.9 & 0.0 & 29.8 & 81 & 82.0 & 18.0 & 0.0& 14.9  & 127 & 34.0 & 66.0 & 0.0\\
		\textbf{MC} & 70 & 82.2 & 0.0 & 17.8 & 61.3 & 81 & 59.4 & 0.0 &40.6& 14.7  & 16 & 78.7 & 0.0 &21.3\\
		\textbf{RW} & 184 & 61.2 & 11.1 & 27.7 & 44.0 & 196 & 6.5 & 5.1 & 88.4& <0.1  & 52 & 41.8 & 31.8 & 26.4
	\end{tabular}
	\medskip

	\caption{Partitions size and learning performance. Discrete cases above bar, continuous below. \textbf{SM}, \textbf{MAN}, \textbf{WW\,1}, \textbf{WW\,2}, \textbf{Nav}, \textbf{BC}, \textbf{MC},  \textbf{RW}, respectively, stand for Simple Maze, Multi-Agent Navigation, Wumpus World\,1, Wumpus World\,2, Navigation, Braking Car, Mountain Car, Random Walk.}%
	\label{tab:tile-cat}

\end{table}

\subsection{RQ1: \Partition\ Size}

\Cref{tab:tile-cat}
shows that SymPar consistently outperforms both tile coding (offline) and CAT-RL (online) on discrete state space cases in terms of success and failure rates, and reduces number of timeouts (\textbf{T}${}_\mathbf{out}$) during learning in majority of cases. 
Also, the agents using SymPar \partitions\ show better performance in terms of
the percentage of episodes during the learning in which they achieve the maximum accumulated reward
in comparison to tile coding partitions (\textbf{Opt}),
cf.\,\cref{tab:tile-cat}.
Note that in
\cref{tab:tile-cat},
the size of \partitions\ is substantially biased in favour of tiling. Nevertheless, SymPar enables better learning. In
\cref{tab:tile-cat},
CAT-RL obtains smaller \partitions\ for \textbf{WW2}, \textbf{MC}, and \textbf{RW} in the same amount of time as SymPar. However, the results for CAT-RL show worse learning performance in comparison to SymPar for these cases as demonstrated by failure and success rates (reporting \textbf{Opt} is not supported by the available CAT-RL implementations, and would require a modification of that method). 
The small partition size in CAT-RL can be explained by its operational mechanism, which involves initialising the agent from a small set. This approach prevents divergence and ensures the number of \parts\ remains constant. Subsequently, CAT-RL implements a policy, aiming to identify the goal state and \partitions\ based on the observations it gathers. Hence, in scenarios where the initial states are not limited and the policies are not goal-oriented, the number of \parts\ will increase. For instance, we have evaluated CAT-RL for mountain car in scenarios where exploration is unrestricted, and the number of \parts\ for a given number of episodes has increased to 302.
For the other test
problems, SymPar achieves better results than CAT-RL in both the partition size and learning performance.
\looseness -1


\begin{figure}[!t]
	\centering
	\includegraphics[width=\columnwidth]{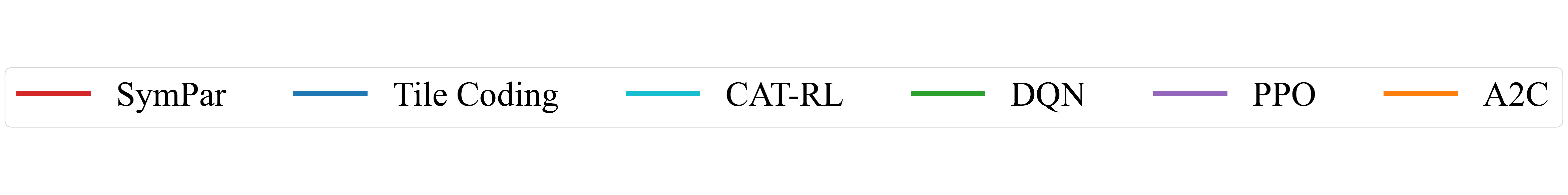}\\
        \subfigure[Braking Car]{
		\includegraphics[width=0.4\columnwidth]{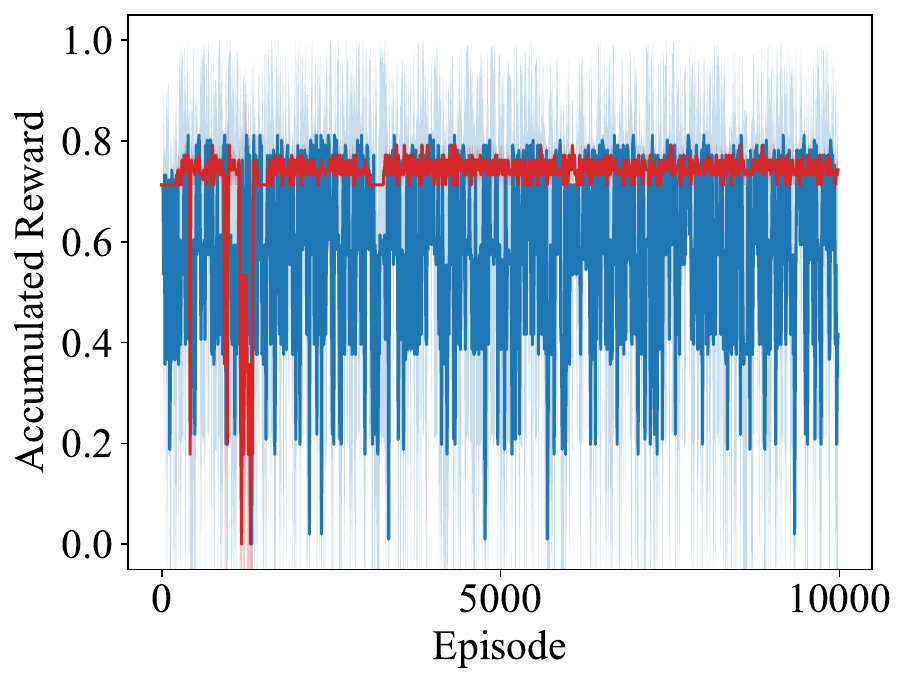}
	}
        \subfigure[Braking Car]{
		\includegraphics[width=0.4\columnwidth]{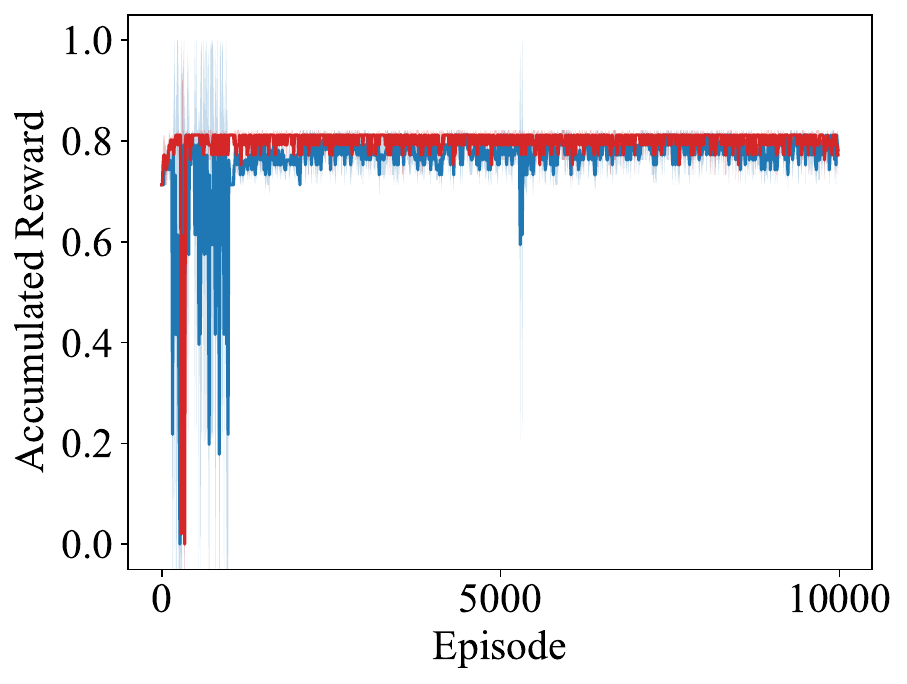}
	}
        \\
        \subfigure[Mountain Car]{
		\includegraphics[width=0.4\columnwidth]{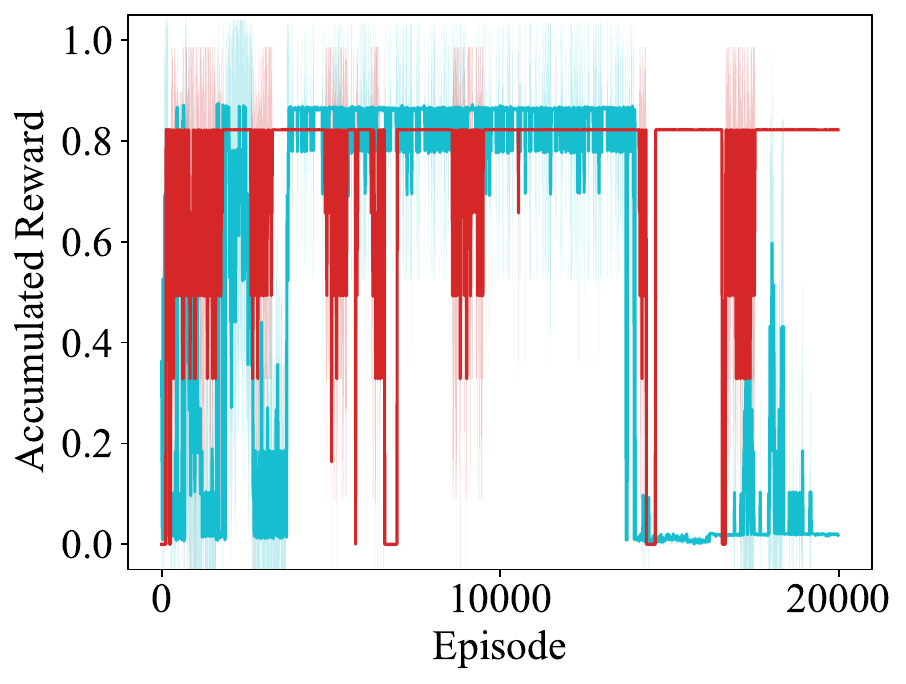}
	}
        \subfigure[Mountain Car]{
		\includegraphics[width=0.4\columnwidth]{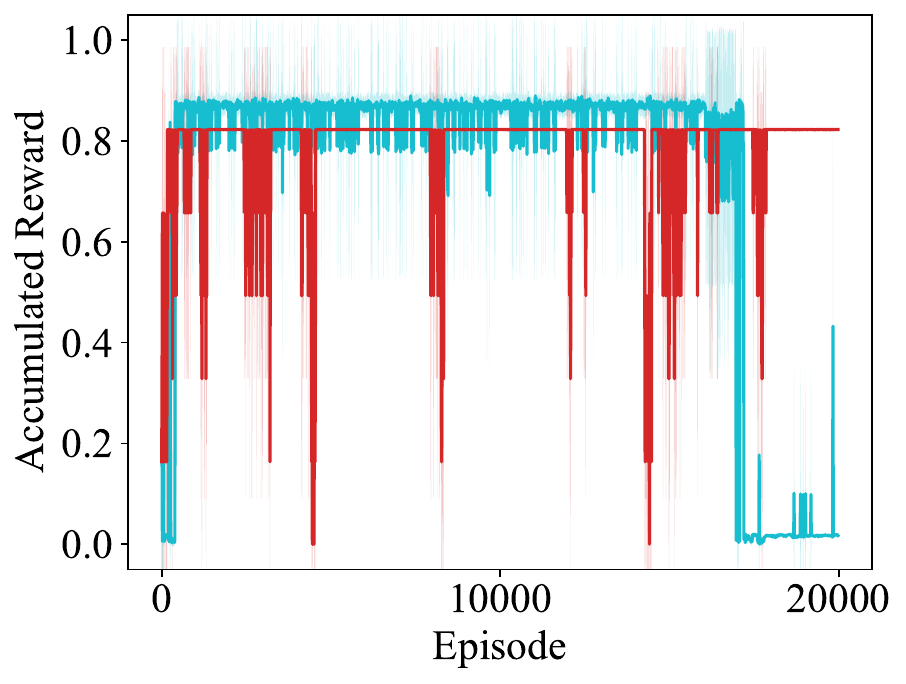}
	}
        \\
        \subfigure[Wumpus World]{
		\includegraphics[width=0.4\columnwidth]{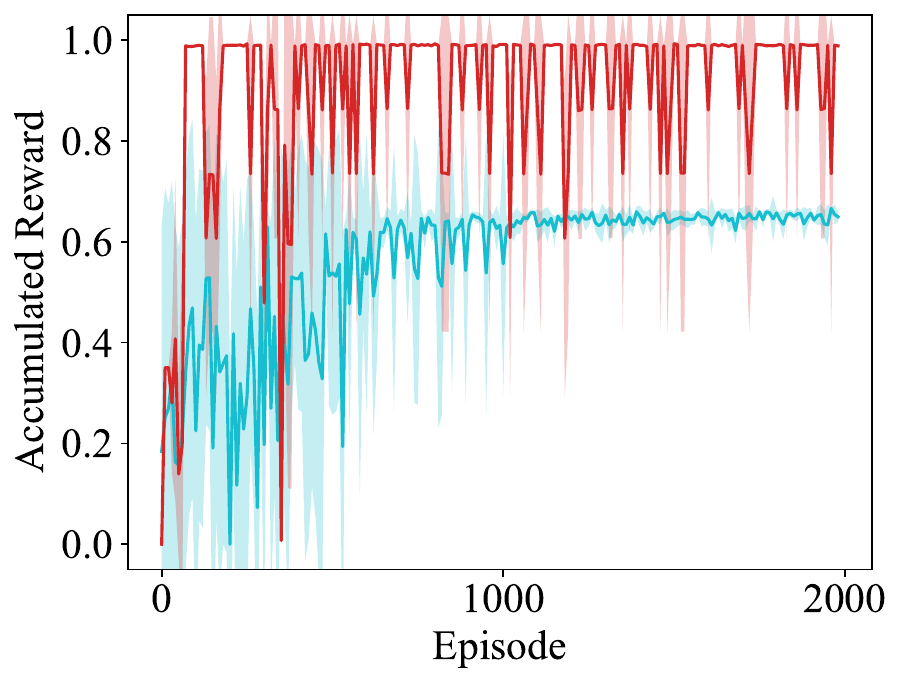}
	}
	\subfigure[Wumpus World]{
		\includegraphics[width=0.4\columnwidth]{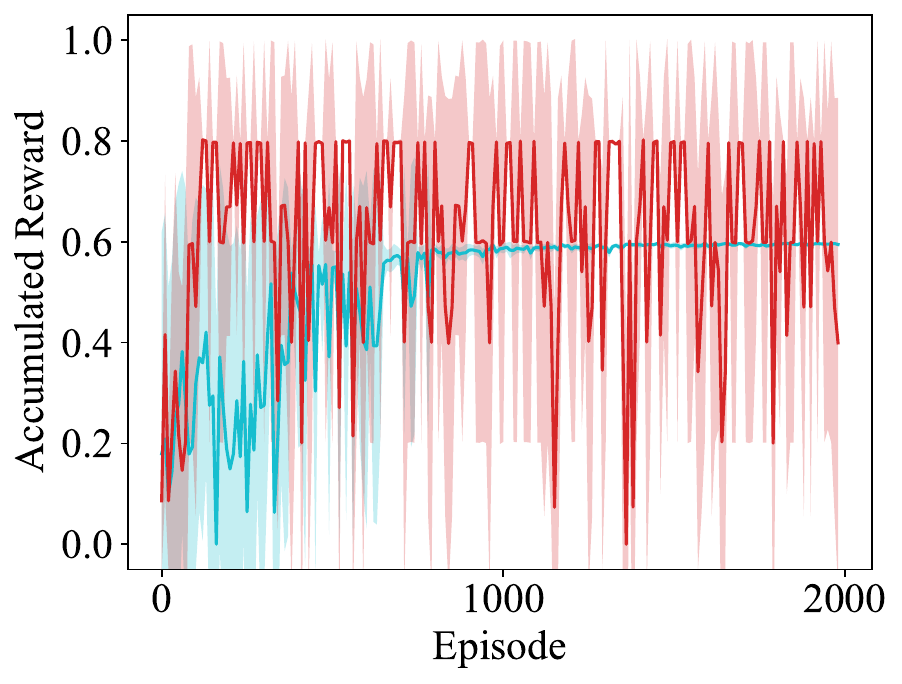}
	}\\
	\caption{Normalized cumulative reward per episode while evaluating ten random states (Left), and less likely states (Right). The best approach for each case is shown.
    }
	\label{fig:accR}


\end{figure}

\begin{figure}[!t]
     \centering
    \subfigure[Braking Car]{
    \resizebox{0.4\columnwidth}{!}{
    \begin{tikzpicture}
    \begin{axis}[
    ybar,
    bar width=0.3cm,
    xlabel={\Huge k},
    axis lines*=left,
	ylabel={\Huge|\EnvState|},
	xticklabel style={font=\LARGE},
	yticklabel style={font=\LARGE},
    ylabel style={yshift=-18pt},
    ymin=-1.1, ymax=1.1,
    ]
    \addplot[blue, fill=blue!30] coordinates {
        (1, 0.01)
        (2, 0.01)
        (3, 0.01)
        (4, 0.06)
        (5, 0.13)
        (6, 0.43)
        (7, 1)
        (8, 1)
        (9, 1)
        (10, 1)
        };
    \end{axis}

    \begin{axis}[
        ybar,
        bar width=0.3cm,
        axis lines*=right,
        ylabel={\Huge Reward of $\pi^*$},
        xtick=\empty,
		yticklabel style={font=\LARGE},
        ylabel style={yshift=10pt},
        ymin=-1.1, ymax=1.1,
        ]
        \addplot[red, fill=red!30] coordinates {
            (1, -1)
            (2, -1)
            (3, -1)
            (4, -0.5)
            (5, -0.5)
            (6, -0.005)
            (7, -0.0001)
            (8, -0.0001)
            (9, -0.0001)
            (10, -0.0001)
        };
    \end{axis}
    \end{tikzpicture}
    }
    \label{fig:k-depthBraking}
    }
    \subfigure[MA Navigation]{
    \resizebox{0.4\columnwidth}{!}{
    \begin{tikzpicture}
        \begin{axis}[
        ybar,
        bar width=0.3cm,
	    xlabel={\Huge k},
        axis lines*=left,
     	xticklabel style={font=\LARGE},
        yticklabel style={font=\LARGE},
		ylabel={\Huge|\EnvState|},
        ylabel style={yshift=-18pt},
        ymin=-1.1, ymax=1.1,
        ]
        \addplot[blue, fill=blue!30] coordinates {
            (1, 0.007)
            (2, 0.007)
            (3, 0.007)
            (4, 0.007)
            (5, 0.007)
            (6, 0.007)
            (7, 0.007)
            (8, 0.007)
            (9, 0.007)
            (10, 0.007)
            (11, 0.007)
            (12, 0.04)
            (13, 0.57)
            (14, 0.94)
            (15, 1)
            };
        \end{axis}

        \begin{axis}[
            ybar,
            bar width=0.3cm,
            axis lines*=right,
	        ylabel={\Huge Reward of $\pi^*$},
	        xtick=\empty,
	        yticklabel style={font=\LARGE},
            ylabel style={yshift=10pt},
            ymin=-1.1, ymax=1.1,
            ]
            \addplot[red, fill=red!30] coordinates {
                (1, -1)
                (2, -1)
                (3, -1)
                (4, -1)
                (5, -1)
                (6, -1)
                (7, -1)
                (8, -1)
                (9, -1)
                (10, -1)
                (11, -1)
                (12, -1)
                (13, -0.022)
                (14, -0.017)
                (15, -0.015)
            };
        \end{axis}
    \end{tikzpicture}
    }
    \label{fig:k-depthMANav}
    }
    \subfigure[Simple Maze]{
    \resizebox{0.4\columnwidth}{!}{
    \begin{tikzpicture}
        \begin{axis}[
    ybar,
    bar width=0.3cm,
    xlabel={\Huge k},
    axis lines*=left,
	ylabel={\Huge|\EnvState|},
	xticklabel style={font=\LARGE},
	yticklabel style={font=\LARGE},
    ylabel style={yshift=-18pt},
    ymin=-1.1, ymax=1.1,
    ]
    \addplot[blue, fill=blue!30] coordinates {
        (1, 0.03)
        (2, 0.03)
        (3, 0.03)
        (4, 0.03)
        (5, 0.03)
        (6, 0.03)
        (7, 0.06)
        (8, 0.12)
        (9, 0.72)
        (10, 0.81)
        (11, 1)
        };
    \end{axis}
    \begin{axis}[
        ybar,
        bar width=0.3cm,
        axis lines*=right,
        ylabel={\Huge Reward of $\pi^*$},
        xtick=\empty,
        yticklabel style={font=\LARGE},
        ylabel style={yshift=10pt},
        ymin=-1.1, ymax=1.1,
        ]
        \addplot[red, fill=red!30] coordinates {
            (1, -1)
            (2, -1)
            (3, -1)
            (4, -1)
            (5, -1)
            (6, -1)
            (7, -1)
            (8, -1)
            (9, -1)
            (10, -0.6)
            (11, -0.4)
        };
    \end{axis}
    \end{tikzpicture}
    }
    \label{fig:k-depthMaze}
    }
    \subfigure[Random Walk]{
    \resizebox{0.4\columnwidth}{!}{
    \begin{tikzpicture}
        \begin{axis}[
    ybar,
    bar width=0.3cm,
    xlabel={\Huge k},
    axis lines*=left,
    xticklabel style={font=\LARGE},
    yticklabel style={font=\LARGE},
    ylabel={\Huge|\EnvState|},
    ylabel style={yshift=-18pt},
    ymin=-1.1, ymax=1.1,
    ]
    \addplot[blue, fill=blue!30] coordinates {
        (1, 0.005)
        (2, 0.005)
        (3, 0.005)
        (4, 0.04)
        (5, 0.09)
        (6, 0.14)
        (7, 0.53)
        (8, 1)
        (9, 1)
        (10, 1)
        };
    \end{axis}

    \begin{axis}[
        ybar,
        bar width=0.3cm,
        axis lines*=right,
        ylabel={\Huge Reward of $\pi^*$},
        xtick=\empty,
        yticklabel style={font=\LARGE},
        ylabel style={yshift=10pt},
        ymin=-1.1, ymax=1.1,
        ]
        \addplot[red, fill=red!30] coordinates {
            (1, -1)
            (2, -1)
            (3, -1)
            (4, -1)
            (5, 0)
            (6, 0)
            (7, 0)
            (8, 0)
            (9, 0)
            (10, 0)
        };
    \end{axis}
    \end{tikzpicture}
    }
    \label{fig:k-depthRandomWalk}
    }
    \caption{Normalized granularity of states and its performance for  symbolic execution with search depth $k$.
      }
    \label{fig:k-depth}

\end{figure}


\begin{figure}[!t]
     \centering
    \subfigure[Braking Car]{
    	\raisebox{-1.5mm}{
	    \includegraphics[width=0.52\columnwidth]{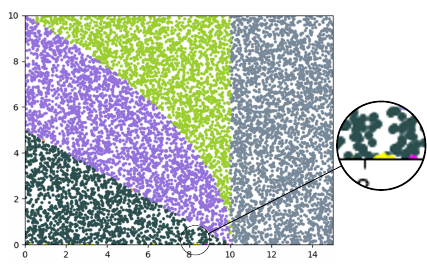}}
    \label{fig:breakingPart}
    }
    \subfigure[Simple Maze]{
		    \includegraphics[width=0.4\columnwidth, height=0.19\textheight]{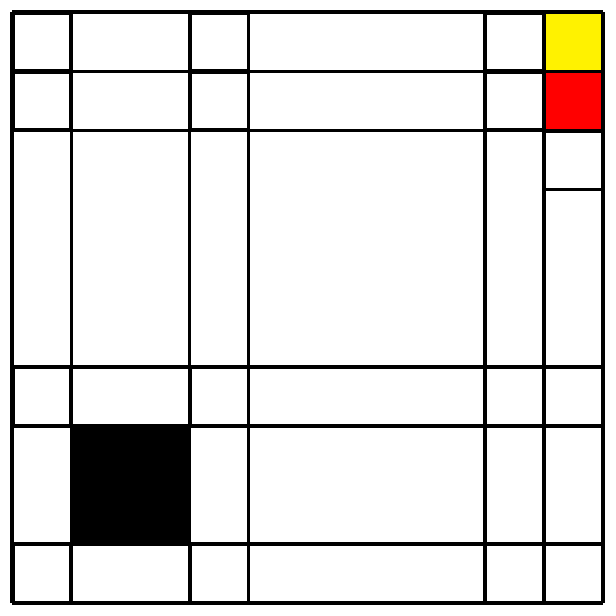}
    \label{fig:maze1010Part}
    }
    \caption{\Partitions\ with SymPar for the Braking Car and Simple Maze.}
    \label{fig:Part}

\end{figure}

For randomly selected states, the three left plots
in \cref{fig:accR}, show that the agents trained by SymPar obtain a
better normalized cumulative reward and subsequently converge faster to a
better policy than the best competing approaches.
The three right plots in the figure show
the accumulated reward when starting from unlikely states (small \parts) for the best competing approaches.
Here, we expect to observe
a good policy from algorithms that capture
the dynamics of environment.  Interestingly, the online technique CAT-RL struggles when dealing with large sets of
initial states.
This can be seen in, e.g., the training for Braking Car,
where each episode introduces new positions and velocities.
\looseness -1

\subsection {RQ2: Granularity vs Learning}

The
plots in
\cref{fig:k-depth} shows that a higher granularity of \partitions\ yields
a higher
accumulated reward
achieved with the optimal policy. 
To be more specific, increasing the depth of search for symbolic execution would result in additional constraints on each \PC, consequently a finer \partition. Then, given sufficient repetition of RL algorithms, finer \partitions\ can yield a better policy for each \parti, due to a reduction in the variance of optimal policies across states in the \parti. This results in a higher accumulated reward when both \partitions\ are evaluated for the same states.

The
plots in \cref{fig:Part} show the shapes of
\partitions\ obtained by SymPar for Braking Car and Simple Maze.  The first plot represents different \parts\ with different colors.  Notably, the green and purple \parts\ depict \partition\ expressions that contain a non-linear relation between the components of the state space (position and velocity).  Besides, close to the x-axis, narrow \parts\ are discernible, depicted in yellow and pink. To illustrate the \partitions\ obtained for Simple Maze, the expressions are translated into a $10\!\times\!10$ grid. The maze used for \cref{fig:maze1010Part} differs from the one before, by including additional obstacles in the environment. These two visualizations shed light on the intricacies of state space \partitioning\ and hint at the logical explainability of the \partitions\ obtained by SymPar. 
\looseness -1

\subsection{RQ3: Scalability}

\Cref{tab:size} shows that the number of parts in SymPar \partitions\ is independent of the size of the state
space. However, this does not imply the universal applicability of the
same \partition\ across different sizes. The conditions specified
within the \partitions\ are size-dependent. Consequently, when analyzing
environments with different sizes for a given problem, running SymPar
is necessary to ensure the appropriate \partition, even though the
total number of \parts\ remains the same.\looseness-1

\subsection{RQ4: \Partitioning\ as an Abstraction}

\Cref{tab:markovity}
presents the variance in accumulated rewards for concrete states across various \parts. The findings demonstrate a notable consistency in accumulated rewards among states within the same \parti, indicating minimal divergence. This is particularly evident when the mean and normalized standard deviation are compared, which demonstrates that the standard deviation is considerably smaller in relation to the mean accumulated reward.


\begin{table}[t!]
	\centering
	\setlength {\tabcolsep} {3pt}
	\begin{tabular}
		{@{\hspace{0pt}}>{\footnotesize}p{10mm}
			>{\footnotesize}l
			>{\footnotesize}c
			@{\hspace{3pt}}>{\footnotesize}p{13mm}
			>{\footnotesize}l
			>{\footnotesize}c
			@{\hspace{3pt}}>{\footnotesize}p{16mm}
			>{\footnotesize}l
			>{\footnotesize}c
		}
		&\footnotesize\bm{{|\EnvState|}}
		&\footnotesize\bm{{|\StateSet{}|}}
		&&\footnotesize\bm{{|\EnvState|}}
		&\footnotesize\bm{{|\StateSet{}|}}
		&&\footnotesize\bm{{|\EnvState|}}
		&\footnotesize\bm{{|\StateSet{}|}}
		\\
		\midrule
		\multirow{3}{1pt}{\textbf{Simple Maze}} &$10\!\times\!10$&33&
		\multirow{3}{1pt}{\textbf{Wumpus World}} & $64\!\times\!64$& 73&
		\multirow{3}{1pt}{\textbf{Navigation}} &$10\!\times\!10$&51\\
		&$10^2\!\times\!10^2$&33&&$10^2\!\times\!10^2$& 73&&$10^2\!\times\!10^2$&51\\
		&$10^3\!\times\!10^3$&33&&$10^3\!\times\!10^3$& 73&&$10^3\!\times\!10^3$&51\\
	\end{tabular}
	\caption{Size of state space and \partition\ for test problems.}%
	\label{tab:size}


\end{table}

\begin{table}[t]
    \centering
    \setlength {\tabcolsep} {2pt}
    
    \begin{tabular}
    {@{\hspace{0pt}}>{\footnotesize}
                      p{10mm}
        >{\footnotesize}c
        >{\footnotesize}c
        >{\footnotesize}c
        >{\footnotesize}c
        >{\footnotesize}c
        }
         &  \bm{$\mathcal{P}_1$}& \bm{$\mathcal{P}_2$} & \bm{$\mathcal{P}_3$} & \bm{$\mathcal{P}_4$} & \bm{$\mathcal{P}_5$}\\
         \midrule
         \textbf{BC} & $-0.05 \pm 0.0\%$ & $-0.01 \pm 0.0\%$ & $-0.5 \pm 0.0\%$ & $-10.0 \pm 0.0\%$ & $ -10.01\pm 0.0\%$ \\
         \textbf{MC} & $996.1 \pm 0.1\%$ & $ 975.5\pm 0.2\%$ & $ 979.04\pm 0.2\%$ & $986.7 \pm 0.2\%$ & $ 981.6\pm 0.2\%$ \\
         \textbf{WW\,1} & $ 486.8\pm 0.3\%$ & $ 490.0\pm 0.2\%$ & $ 477.6\pm 0.2\% $ & $ 475.0\pm 0.0\%$ & $ 495.8\pm 0.1\% $ \\
    \end{tabular}
    \medskip
    \caption{Assessment of similarity of concrete states within \parts. \textbf{BC}, \textbf{MC}, \textbf{WW\,1}, respectively, stand for Braking Car, Mountain Car, Wumpus World\,1.}%
    \label{tab:markovity}
\end{table}

\begin{table}[t!]
\centering
\setlength {\tabcolsep} {3.0pt}
\begin{tabular}
	{@{\hspace{0pt}}>{\footnotesize}p{15mm}
		>{\scriptsize}c
		>{\scriptsize}c
		>{\scriptsize}c
		>{\scriptsize}c
		>{\scriptsize}c
		>{\scriptsize}c
	}
	&
 \footnotesize\textbf{Off}&  \footnotesize\textbf{Auto}&
 \footnotesize \textbf{Dyn}& \footnotesize\textbf{NonL}&
 \footnotesize\textbf{NarrP}&
 \footnotesize\textbf{SInd}\\
	\midrule
	\textbf{SymPar}&  \checkmark&  \checkmark&  \checkmark&  \checkmark&\checkmark&\checkmark\\
	\textbf{CAT-RL}&  $\times$&  \checkmark& \checkmark & $\times$ &$\times$&$\times$\\
	\textbf{Tiling}& \checkmark & $\times$ & $\times$& $\times$ &$\times$&$\times$\\
\end{tabular}

\smallskip

\caption{Capabilities and properties.}%
\label{tab:prop}


\end{table}

\paragraph{Summary.}
Our experiments show distinct advantages of SymPar over the
other approaches, cf.\ \cref{tab:prop}. It is an
offline (\textbf{Off}) automated (\textbf{Auto}) approach, which
captures the dynamics of the environment (\textbf{Dyn}), and maps the
nonlinear relation between components of the state into their
representation (\textbf{NonL}). SymPar can detect narrow
\parts\ (\textbf{NarrP}) without excessive sampling
and generates a logical \partition\ that is independent of the
specific size of the state space (\textbf{SInd}). This comprehensive comparison
underscores the robust capabilities of SymPar across various
dimensions, positioning it as a versatile and powerful approach
compared to CAT-RL and Tile Coding.
\looseness -1

\paragraph {Threats to Internal Validity.}

The data produced in response to RQ1 and RQ2 may be incorrectly interpreted  as suggesting existence of a correlation between the size of partitions and the effectiveness of learning.  No such obvious correlation exists: too small, too large, and incorrectly selected partitions hamper learning.  We cannot claim any such correlation. We merely report the size of the partitions and the performance of learning for the selected cases.

While we study the impact of the state space size on SymPar (RQ3), one should remember that there is no strong relationship between the size and the complexity of the state space. In general, the complexity (the branching of the environment model) has a dominant effect on the performance of symbolic execution.
\looseness -1

We assumed that two states are similar if they have yield similar accumulated reward in the obtained policy (RQ4).  First, note that we have no guarantee that the used policy is optimal, although the plots suggest convergence. Second, a more precise, but also more expensive, alternative would be to compute the optimal policy for each of the two states separately (taking them out of partitions). This could lead to higher and different reward values. However, even this would not guarantee the reliability of the estimated state values, as the hypothetical optimal accumulated reward requires representing the optimal policy precisely for all reachable states, which is infeasible in a continuous state space with a probabilistic environment.

\paragraph {Threats to External Validity.}

The results of experiments are inherently not generalizable, given that we use a finite set of cases. However, the selected cases do cover a range of situations: discrete and continuous state spaces, deterministic and non-deterministic environments, as well as single- and multi-agent environments.

\paragraph{Technical Details.}

The implementation of SymPar (will be publicly available upon the acceptance) uses Symbolic
PathFinder\footnote{https://github.com/SymbolicPathFinder}
\cite{pasareanu13ase} as its symbolic executor,
Z3\footnote{https://github.com/Z3Prover/z3} \cite{moura08tacas} as its
main SMT-Solver and
the SMT-solver
DReal\footnote{https://github.com/dreal/dreal4} \cite{gao13cade}
to handle non-linear functions such as trigonometric
functions.

\looseness -1


\section{Discussion and Limitations}%
\label{sec:Discussion}

SymPar is not limited to reinforcement learning. Theoretically, it could be applied with traditional solving techniques for MDPs. However, this would require efficient methods for extracting MDP models from simulator code.

SymPar uses environments that are implemented as programs, so they are formally specified. This may suggest that one can obtain the policies analytically, not through (statistical) RL. However, many problems exist which we can formulate as programs, but those semantics are too complex to handle with precise analytical methods by solving the derived MDPs. For example, consider an autonomous drone delivery system in an urban setting that needs to transport packages efficiently, while avoiding static (buildings) and dynamic (other drones) obstacles. The urban environment can be modeled with precise geometry and established laws of physics that govern drone flight dynamics. Weather predictions and obstacle patterns can frequently be pre-simulated. Despite the availability of an exact environment model, reinforcement learning remains the preferred solution due to its ability to scale to this complexity, which analytical methods cannot \cite{jevtic2023reinforcement,chen2022deliversense,kretchmara2001robust}.
\looseness -1

Simulations often rely on simplifying assumptions to make them computationally feasible. If these assumptions abstract from critical details, the simulations might not be fully transparent or interpretable. Even in these cases, the efficacy of SymPar in achieving effective \partitions, and the capacity of RL to identify optimal policies, remains valid. While we did not undertake a direct experiment for this scenario, the experiments for RQ2 can serve as a surrogate evidence. Limiting the depth of search for symbolic execution may generate a coarser \partition\ than from a fully analyzed the program, which while not exactly the same, is similar to using a more abstract program for partitioning. These experiments indicate that if a simplified model of the environment is used, SymPar could still generate a \partition\ that can be used for more realistic environment models.

In concurrency theory, lumping or bisimulation minimization is sometimes used as a partitioning technique. Note that bisimulation minimization is presently not possible for environment models expressed as computer programs.  We would need symbolic bisimulation-minimization methods.  Also, note that bisimulation induces a finer partitioning than we need: it puts in a single equivalence class all states that are externally indistinguishable, while we only need to unify states that share the same optimal action in one step. In contrast, symbolic execution performs a mixed syntactic-semantic decomposition of the input state space by means of path conditions. This process is mainly driven by the syntax of the program, yet it is semantically informed via the branch conditions. The obtained \partition\ might be unsound from the bisimulation perspective, but it tends to produce coarser \partitions.
\looseness -1

SymPar analyzes single step executions of the environment. There are however problems where the interesting behaviors are observed only over a sequence of decisions. For example, the dynamics of Cart Pole\,\cite{sutton.barto:2018}  is described by a continuous formula over its position and velocity along with the angle and angular velocity of the pole. There is, in fact, no interesting explicit branching---the path conditions found by symbolic execution are trivial. SymPar is better suited for problems with explicit branching in the environment dynamics. At the same time, excessive branching can hamper its efficiency. In these cases, choosing a reasonable depth may achieve a \partition\ that is sufficiently good while controlling its size (\cref{fig:k-depth}).
\looseness -1


\section{Conclusion}%
\label{sec:Conclusion}

SymPar is a new generic and automatic offline method for \partitioning\ state spaces in reinforcement learning based on a symbolic analysis of the environment's dynamics. In contrast to related work, SymPar's \partitions\ effectively capture the semantics of the environment. SymPar accommodates non-linear environmental behaviors by using adaptive \partition\ shapes, instead of rectangular tiles. Our experiments demonstrate that SymPar improves state space coverage with respect to environmental behavior and allows reinforcement learning to better handle with sparse rewards. However, since SymPar analyzes the simulator of the environment, it is sensitive to the implementation of the environment model. The performance of the underlying tools, including the symbolic executor and SMT solvers, also affect the effectiveness of SymPar for complex simulators with long execution paths.  In the future, we would like to address these limitations and consider using symbolic execution also for online \partitioning.  \looseness -1




\paragraph{Acknowledgment.}
This work was partially funded by DIREC (Digital
Research Centre Denmark), a collaboration between the eight Danish universities
and the Alexandra Institute supported by the Innovation Fund Denmark.

\paragraph{Data Availability Statement.}
The source code of SymPar, the benchmark items, the evaluation results and instructions for reproduction are available online via DOI \url{10.5281/zenodo.14620119}.

\bibliographystyle {splncs04}
\bibliography {references}

 \appendix
 \section {Appendix / Supplementary Material}%
 \label {appendix}

\subsection{Proofs of Properties of SymPar}
\label{sec:proof}
\setcounter{theorem}{0}
\begin{theorem}
	The set  \( \mathcal P \)obtained in \cref{alg:sympar} is a partition (i.e., it is total):
		\(\forall \EnvS \in \EnvState \; \exists!\, \mathcal{P}_0 \in\mathcal{P}\cdot \; \EnvS \in \mathcal{P}_0.\) 
\end{theorem}

\begin{proof}
  The theorem follows from the fact that the partitioning generated by
  SymPar is obtained from first running the simulated environment
  symbolically and collecting the path conditions. By design, all path
  conditions produced by a complete terminating symbolic execution run
  is a partitioning. The final partitioning is obtained by
intersecting these partitionings to obtain the unique coarsest
partitioning finer than each of them. This partitioning is known from
order theory \cite{davey1990introduction} to be unique and it is
total and pairwise disjoint. Hence, it can be inferred that each
concrete state in the partitioned state space, $\EnvS \in \EnvState$,
is represented by at least one partition
$\mathcal{P}_0\in \mathcal{P}$.\looseness -1
\end{proof}

\begin {theorem}
Let \( \PC ^a \) be the set of path conditions produced by SymPar for
each of the actions \( a \in \Action \). The size of the final
\partition\ \Pcal returned by SymPar is bounded from below by each
\(|\PC ^a|\) and from above by \( \prod_{a\in\Action}|\PC^a| \).
\end {theorem}

\begin{proof}
  The theorem follows from the fact that \Pcal is finer than any of
  the \( \PC ^a \)s and the algorithm for computing the coarsest
  partitioning finer than a set of partitionings can in the worst case
  intersect each partition in each set \( PC^a \) with all the
  partitions in the partitionings of the other actions.
\end{proof}

\subsection{Additional Results}
\label{sec:addres}

To make the comparison with
DQN, A2C and PPO fair, we used
the same running time as for SymPar, which resulted in lower
performance for these approaches.
The fluctuation observed in the plots suggest that they may need more
iterations and possibly more customized architectures. A2C and PPO,
which are proper for problems with continuous action space, behave as
expected.

\begin{figure}[!t]
	\centering
	\vspace{-4mm}
	\includegraphics[width=\columnwidth]{Figures/LABELS.pdf}\vspace{-3mm}\\
	\subfigure[Braking Car]{
		\includegraphics[width=0.3\columnwidth]{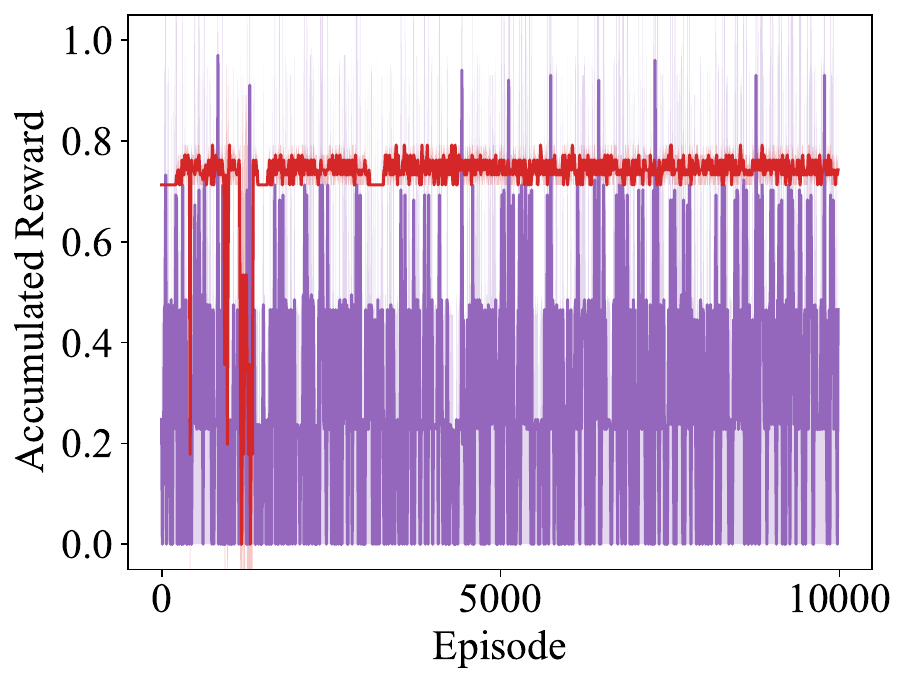}
	}
	\hspace{-3mm}
        \subfigure[Mountain Car]{
		\includegraphics[width=0.3\columnwidth]{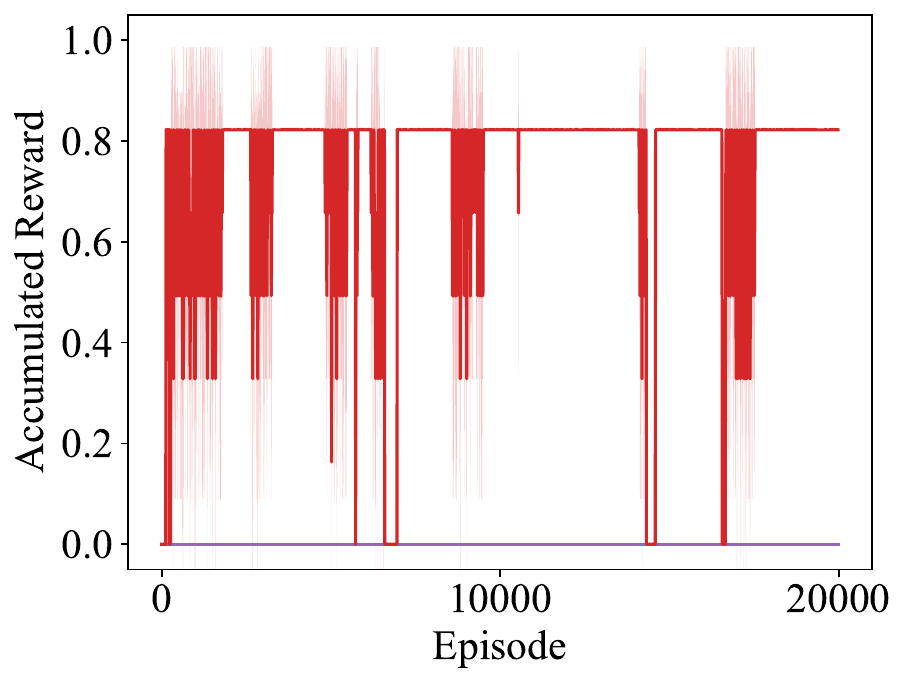}
	}
        \subfigure[Wumpus World]{
		\includegraphics[width=0.3\columnwidth]{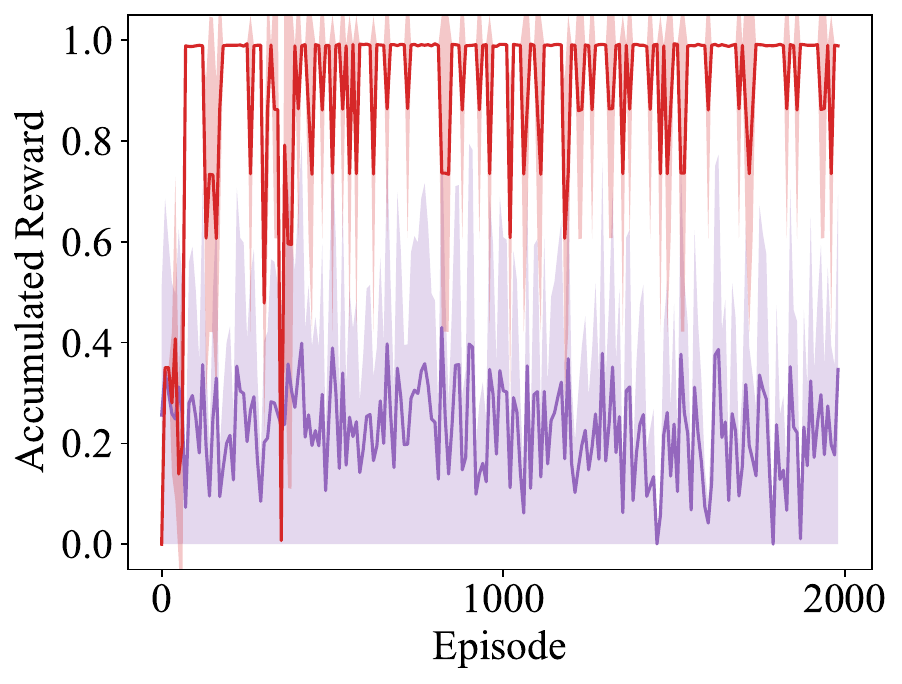}
	}
	\\
	\vspace{-1mm}
	\hspace{-3mm}
	\subfigure[Braking Car]{
		\includegraphics[width=0.3\columnwidth]{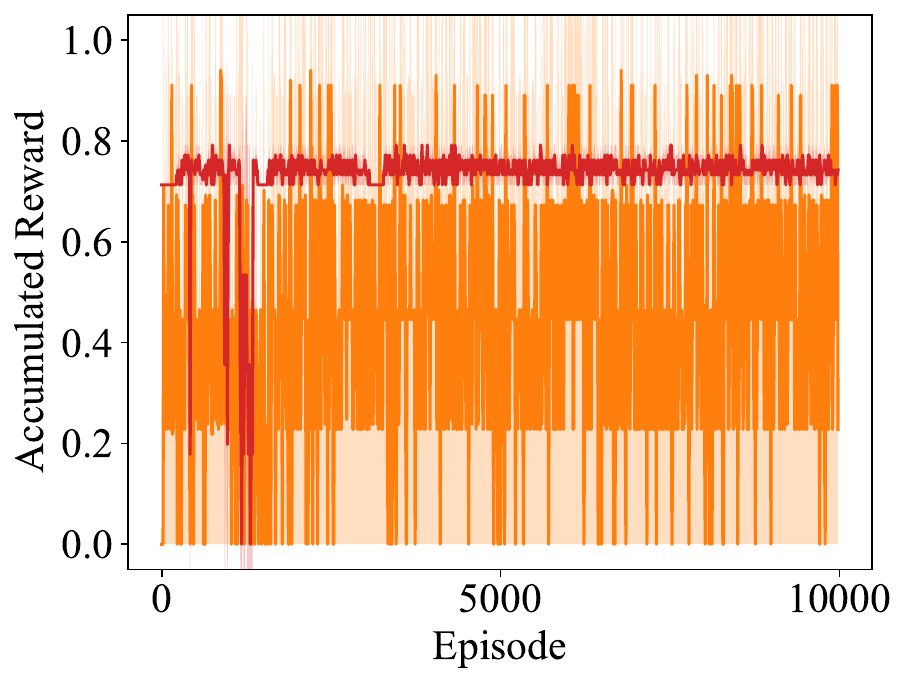}
	}
	\hspace{-3mm}
	\subfigure[Mountain Car]{
		\includegraphics[width=0.3\columnwidth]{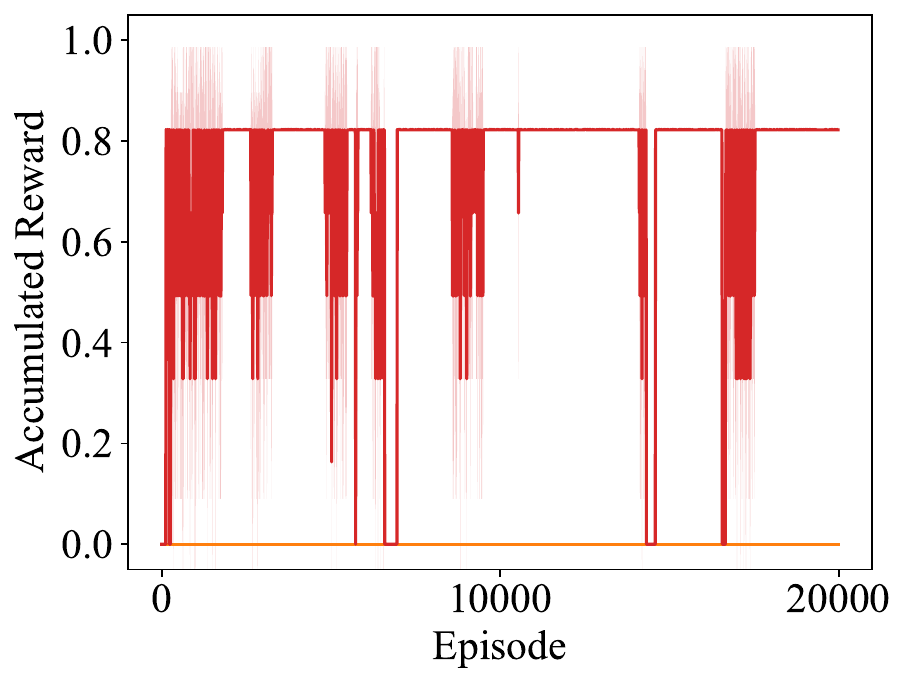}
	}
	\hspace{-3mm}
	\hspace{-3mm}
	\subfigure[Wumpus World]{
		\includegraphics[width=0.3\columnwidth]{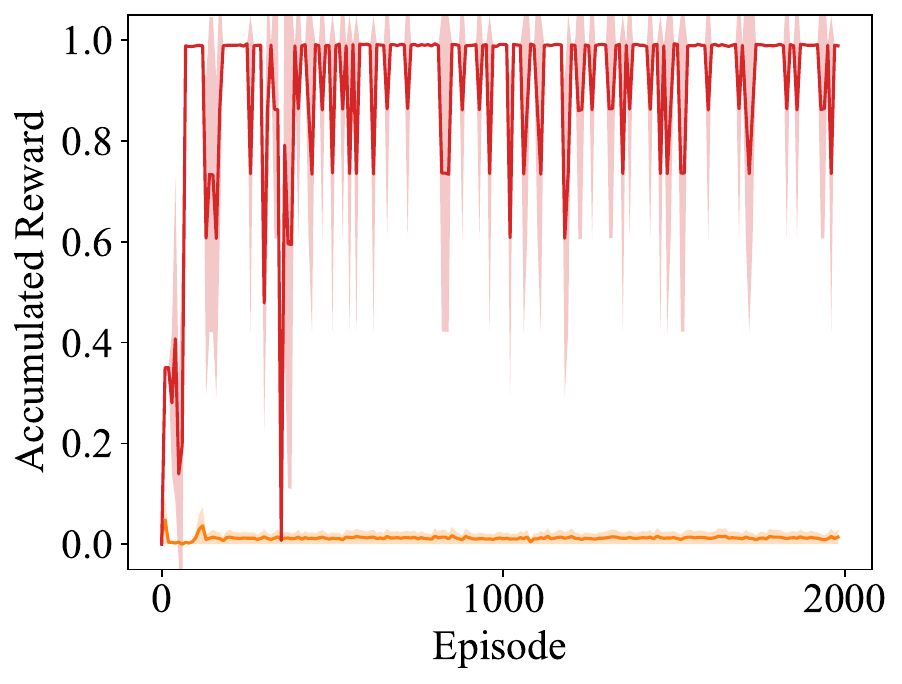}
	}\\
        \vspace{-1mm}
	\hspace{-3mm}
	\subfigure[Braking Car]{
		\includegraphics[width=0.3\columnwidth]{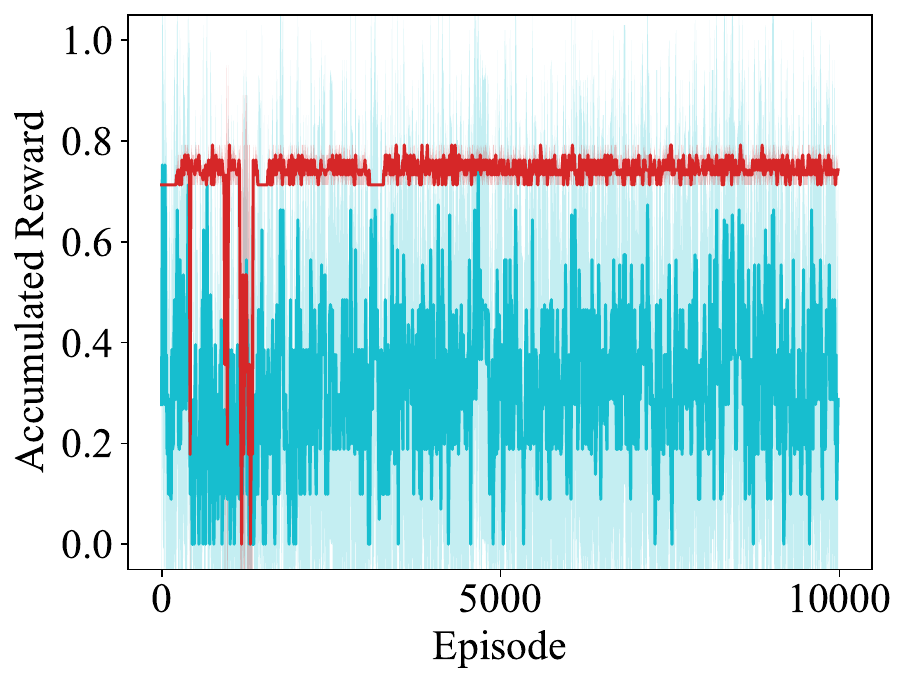}
	}
	\hspace{-3mm}
	\subfigure[Mountain Car]{
		\includegraphics[width=0.3\columnwidth]{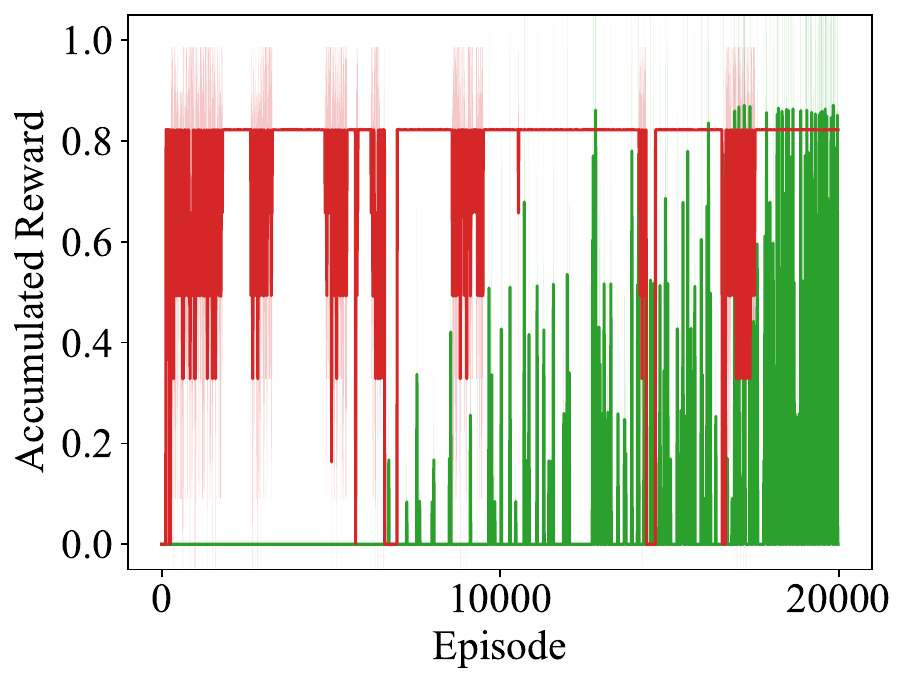}
	}
	\hspace{-3mm}
	\subfigure[Wumpus World]{
		\includegraphics[width=0.3\columnwidth]{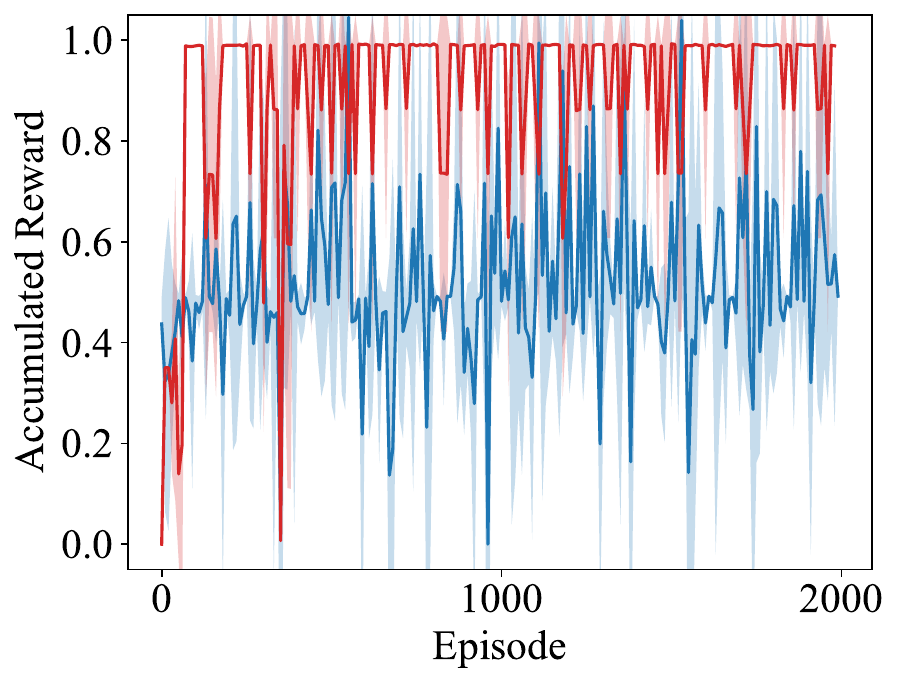}
	}\\
        \vspace{-1mm}
	\hspace{-3mm}
	\subfigure[Braking Car]{
		\includegraphics[width=0.3\columnwidth]{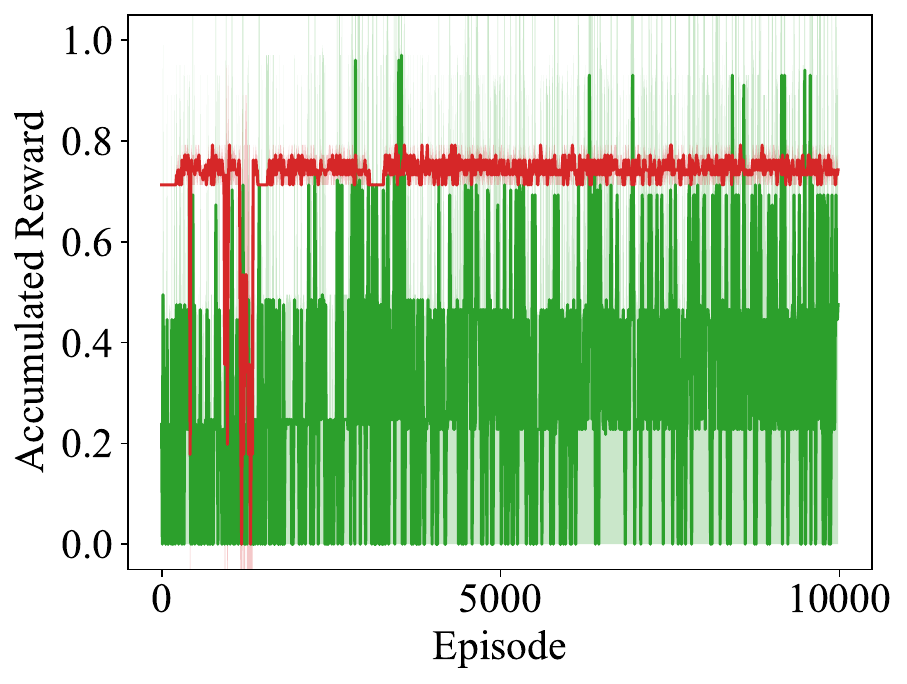}
	}
	\hspace{-3mm}
	\subfigure[Mountain Car]{
		\includegraphics[width=0.3\columnwidth]{Figures/accR-mountaincar2_dqn.pdf}
	}
	\hspace{-3mm}
	\subfigure[Wumpus World]{
		\includegraphics[width=0.3\columnwidth]{Figures/accR-wumpus2_tiling.pdf}
	}\\
	\vspace{-1mm}
	\caption{Normalized cumulative reward per each episode while evaluating ten random states.\vspace{\generalspace}}
	\label{fig:accR2}
\end{figure}

\begin{figure}[!t]
	\centering
	\vspace{-4mm}
	\includegraphics[width=\columnwidth]{Figures/LABELS.pdf}\vspace{-3mm}\\
	\subfigure[Braking Car]{
		\includegraphics[width=0.3\columnwidth]{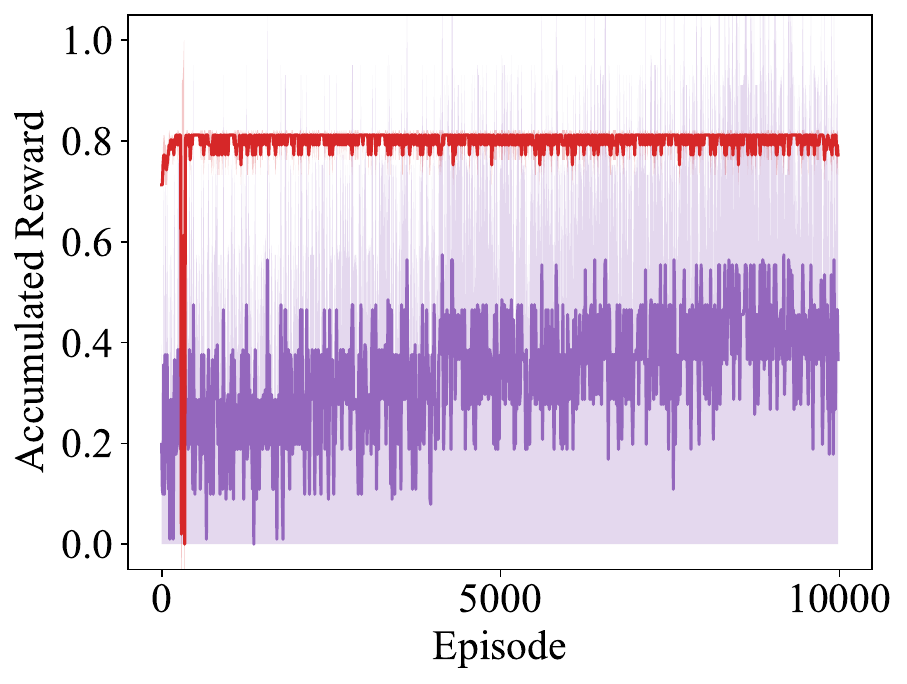}
	}
	\hspace{-3mm}
        \subfigure[Mountain Car]{
		\includegraphics[width=0.3\columnwidth]{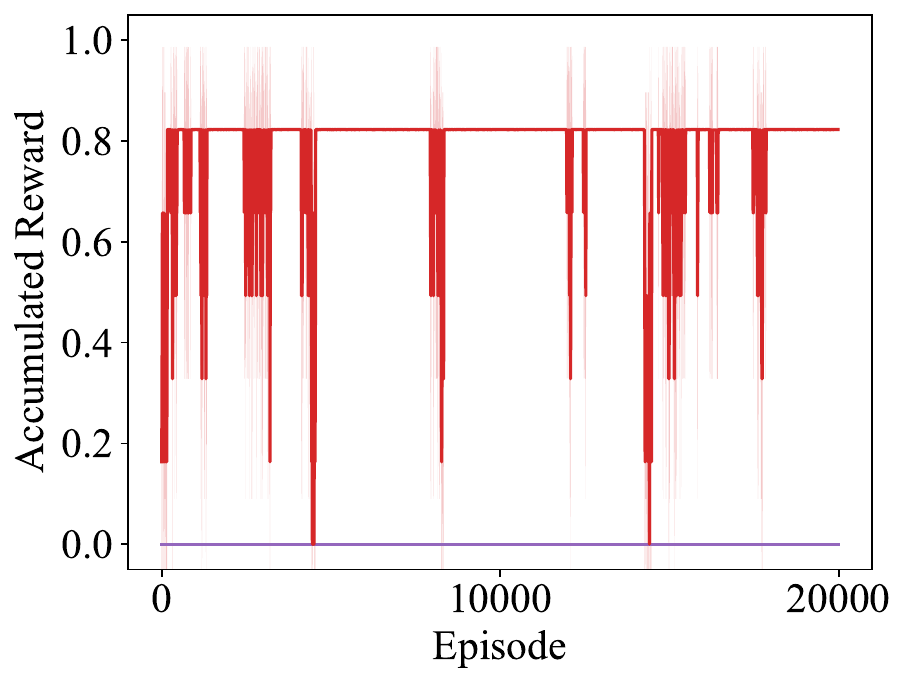}
	}
        \subfigure[Wumpus World]{
		\includegraphics[width=0.3\columnwidth]{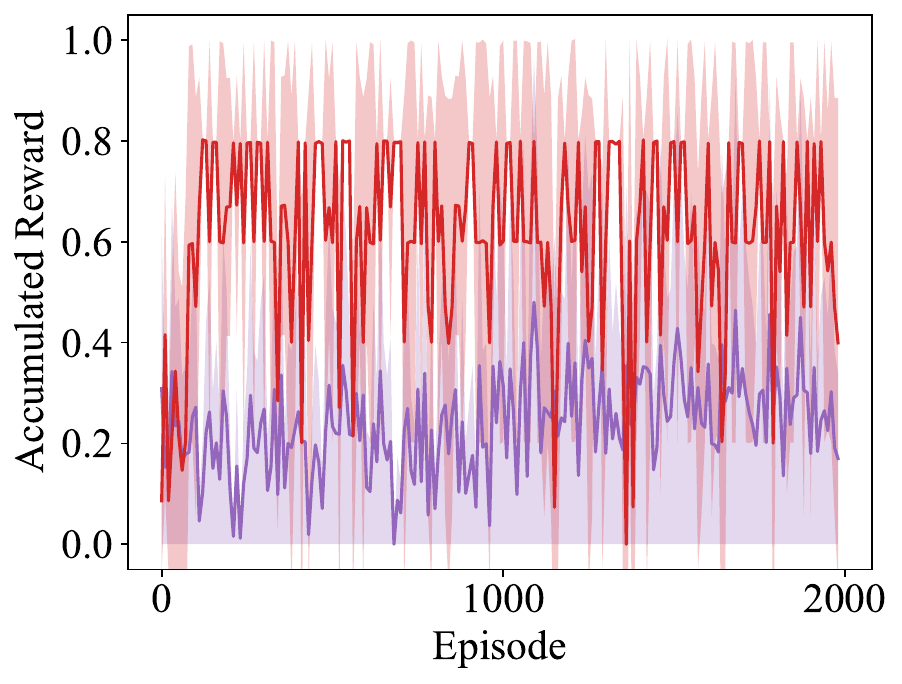}
	}
	\\
	\vspace{-1mm}
	\hspace{-3mm}
	\subfigure[Braking Car]{
		\includegraphics[width=0.3\columnwidth]{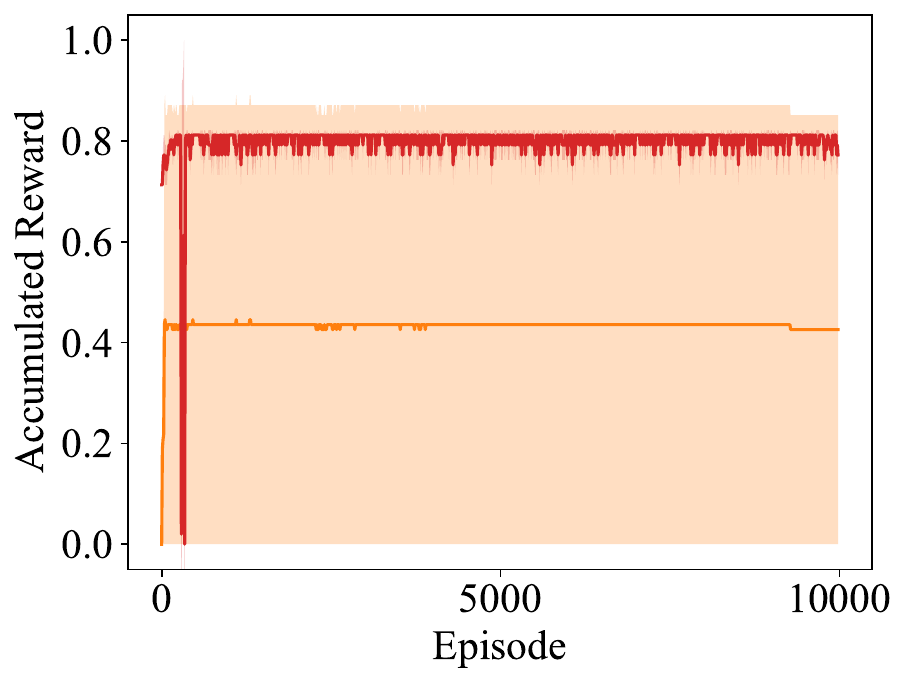}
	}
	\hspace{-3mm}
	\subfigure[Mountain Car]{
		\includegraphics[width=0.3\columnwidth]{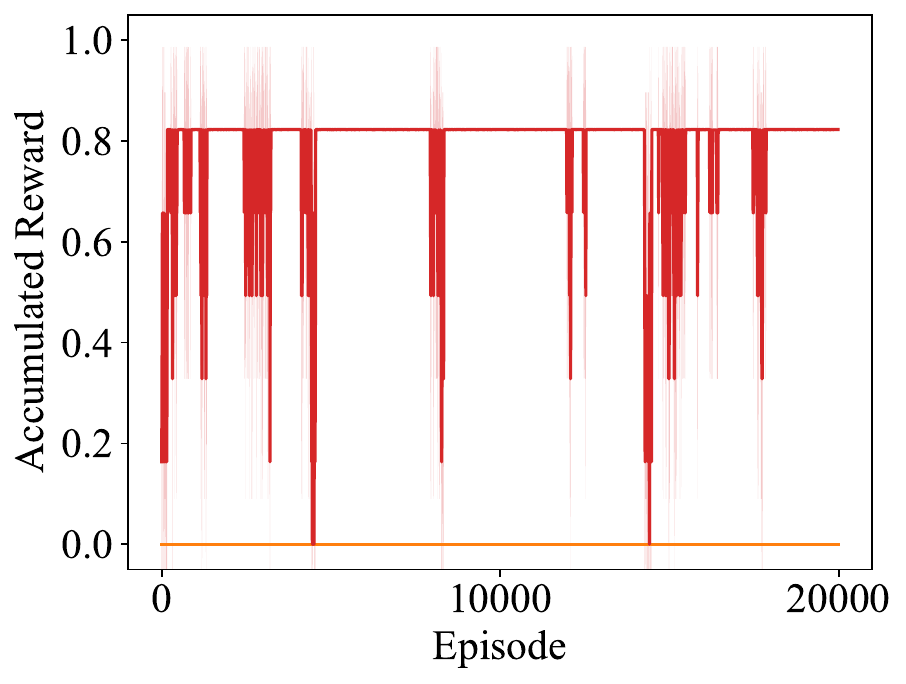}
	}
	\hspace{-3mm}
	\subfigure[Wumpus World]{
		\includegraphics[width=0.3\columnwidth]{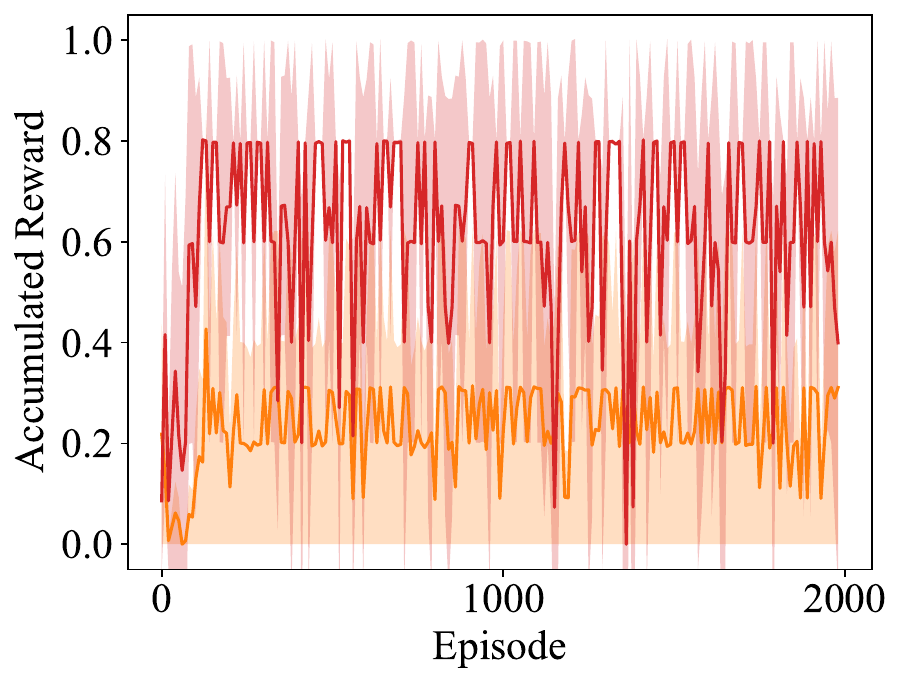}
	}\\
        \vspace{-1mm}
	\hspace{-3mm}
	\subfigure[Braking Car]{
		\includegraphics[width=0.3\columnwidth]{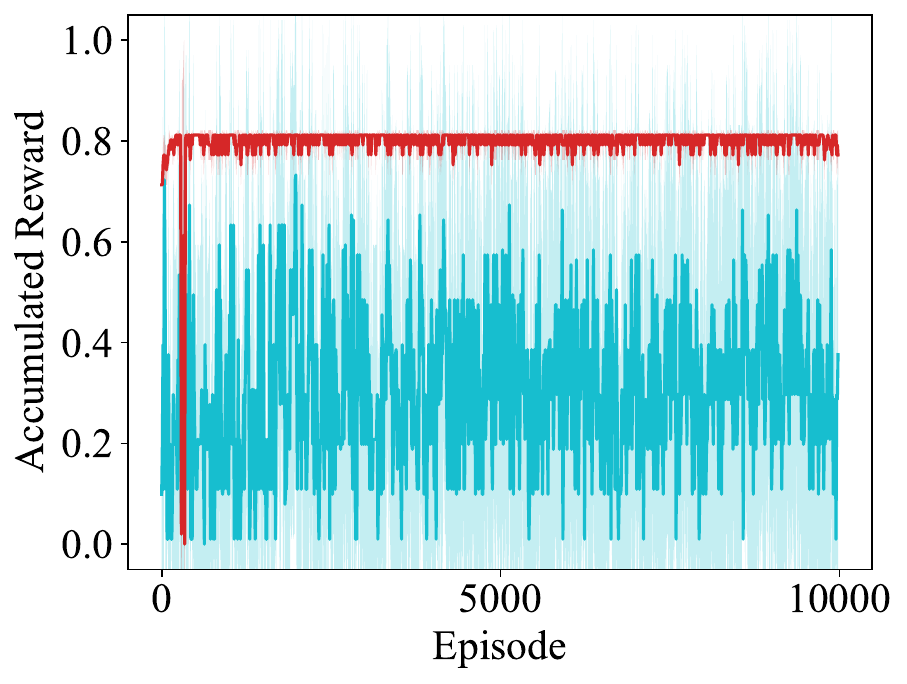}
	}
	\hspace{-3mm}
	\subfigure[Mountain Car]{
		\includegraphics[width=0.3\columnwidth]{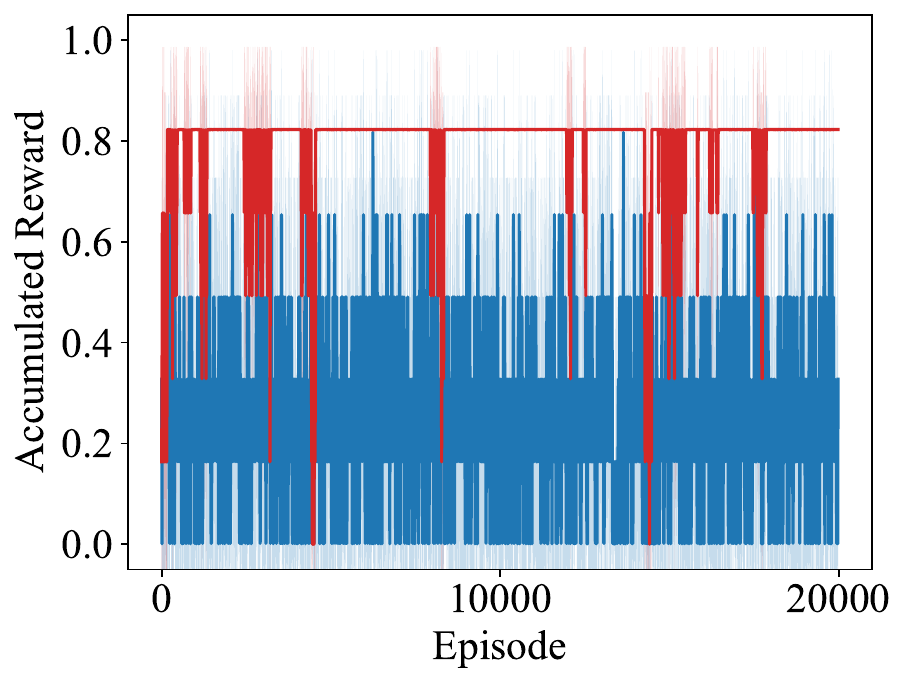}
	}
	\hspace{-3mm}
	\subfigure[Wumpus World]{
		\includegraphics[width=0.3\columnwidth]{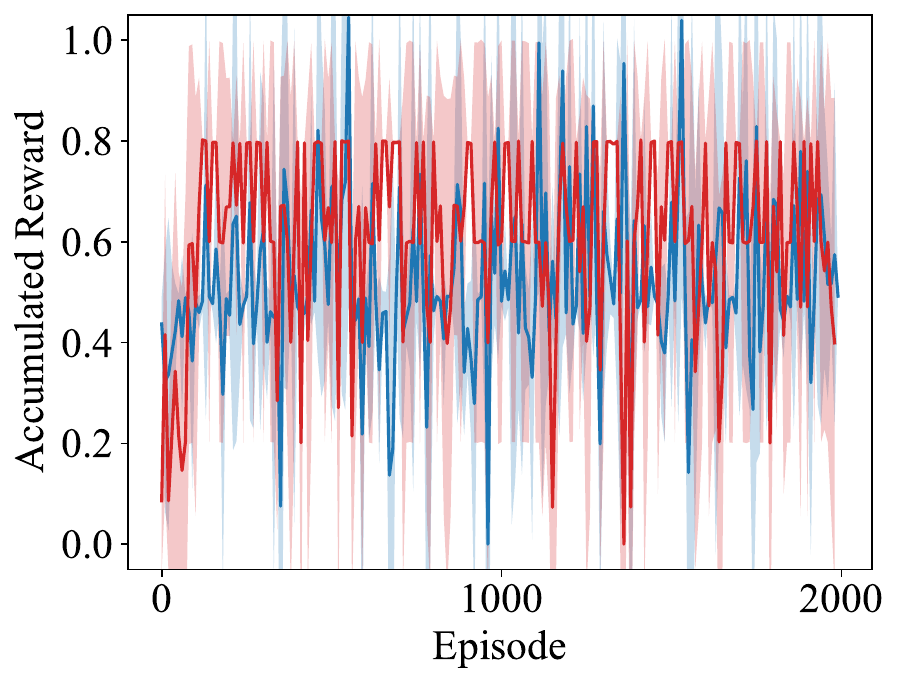}
	}\\
        \vspace{-1mm}
	\hspace{-3mm}
	\subfigure[Braking Car]{
		\includegraphics[width=0.3\columnwidth]{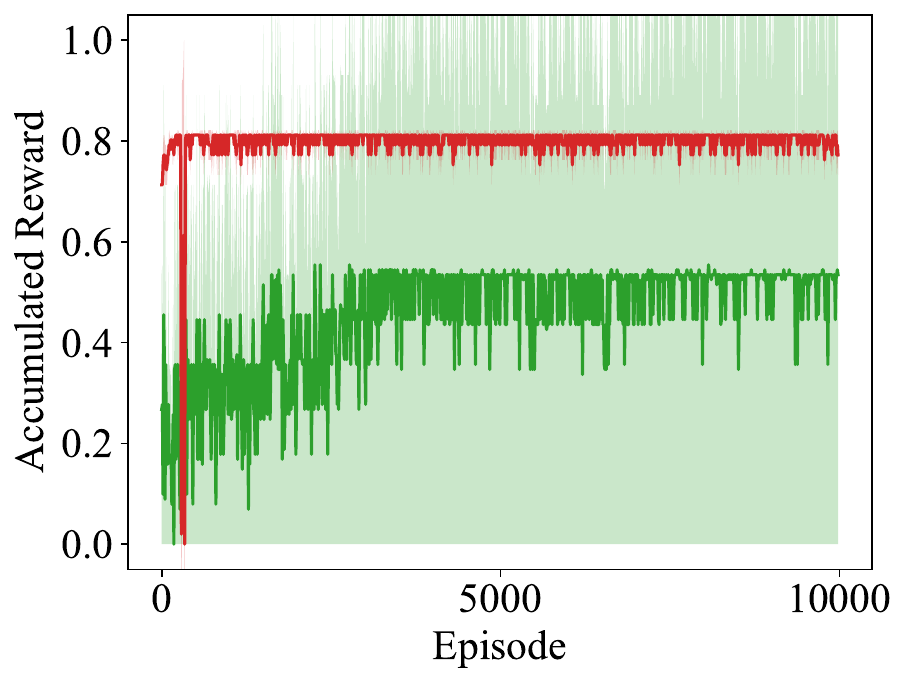}
	}
	\hspace{-3mm}
	\subfigure[Mountain Car]{
		\includegraphics[width=0.3\columnwidth]{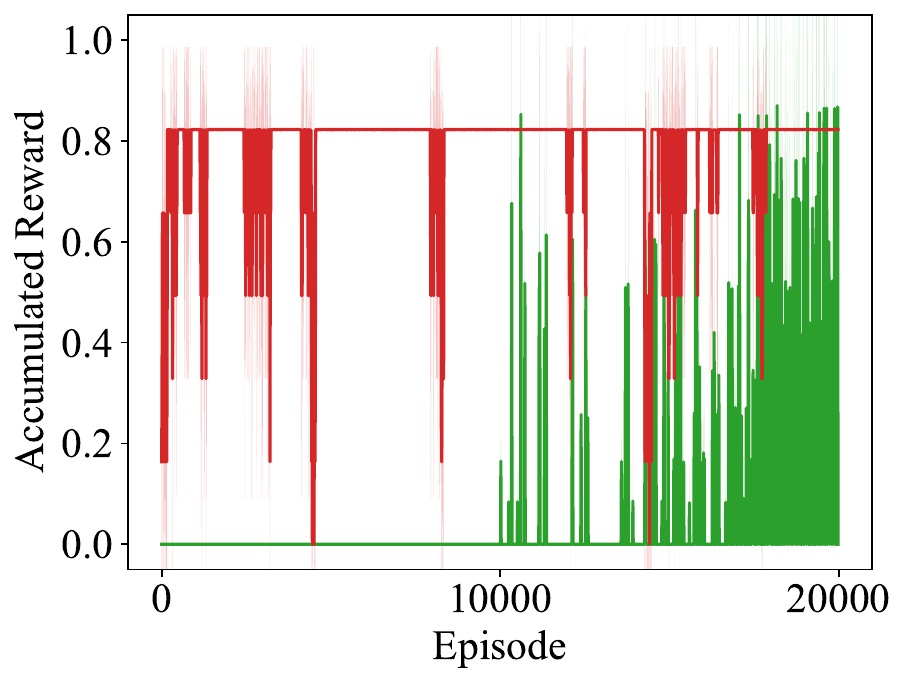}
	}
	\hspace{-3mm}
	\subfigure[Wumpus World]{
		\includegraphics[width=0.3\columnwidth]{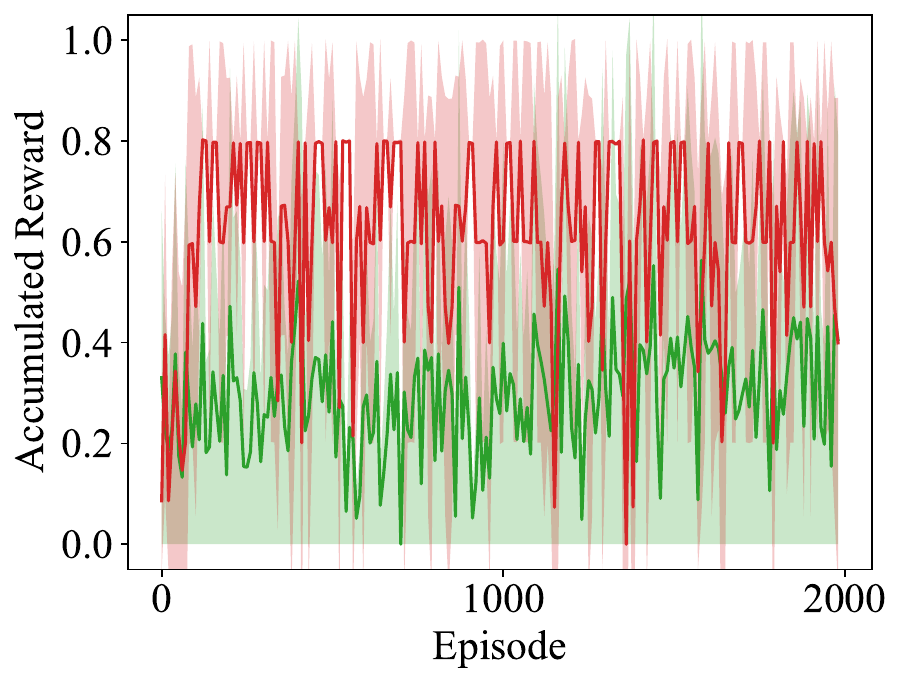}
	}\\
	\vspace{-1mm}
	\caption{Normalized cumulative reward per each episode while evaluating ten less likely states.\vspace{\generalspace}}
	\label{fig:accR4}
\end{figure}

\end {document}